 \documentclass[manuscript,screen]{acmart}
\AtBeginDocument{%
  }
\usepackage{afterpage}
\usepackage{longtable}
\usepackage{booktabs}
\usepackage{array}
\usepackage{makecell}
\usepackage{pifont}
\usepackage{multirow}
\newcommand{\cmark}{\checkmark}        
\newcommand{\xmark}{\ding{55}}        
\newcommand{\pmark}{\dag}              

\usepackage{tikz,xcolor,hyperref}

\acmISBN{978-1-4503-XXXX-X/2018/06}

\begin{document}

\title{Explainable Artificial Intelligence in Biomedical Image Analysis: A Comprehensive Survey}

\author{Getamesay Haile Dagnaw} \orcid{0000-0003-4580-3133}
\author{ Yanming Zhu} \orcid{0000-0002-8238-8090}
\author{ Muhammad Hassan Maqsood} \orcid{0000-0002-3704-1899}
\email{{getamesay.dagnaw, yanming.zhu, muhammadhassan.maqsood}@griffithuni.edu.au}
\affiliation{%
  \institution{Griffith University}
  \city{Gold Coast}
  \state{Queensland}
  \country{Australia}
}

\author{Wencheng Yang}
\email{wencheng.yang@unisq.edu.au}
\orcid{0000-0001-7800-2215}
\affiliation{
\institution{University of Southern Queensland}
\city{Toowoomba} 
\state{QLD}
\country{Australia}}

\author{Xingshuai Dong} \orcid{0000-0003-3900-1038}
\author{ Xuefei Yin} \orcid{0000-0002-5784-7419}
\author {Alan Wee-Chung Liew} \authornotemark[1] \orcid{0000-0001-6718-7584} 
\email{{xingshuai.dong, x.yin, a.liew}@griffith.edu.au}
\affiliation{%
  \institution{Griffith University}
  \city{Gold Coast}
  \state{QLD}
  \country{Australia}
}

\renewcommand{\shortauthors}{Getamesay et al.}

\begin{abstract}
 Explainable artificial intelligence (XAI) has become increasingly important in biomedical image analysis to promote transparency, trust, and clinical adoption of DL models. While several surveys have reviewed XAI techniques, they often lack a modality-aware perspective, overlook recent advances in multimodal and vision-language paradigms, and provide limited practical guidance. This survey addresses this gap through a comprehensive and structured synthesis of XAI methods tailored to biomedical image analysis. We systematically categorize XAI methods, analyzing their underlying principles, strengths, and limitations within biomedical contexts. A modality-centered taxonomy is proposed to align XAI methods with specific imaging types, highlighting the distinct interpretability challenges across modalities. We further examine the emerging role of multimodal learning and vision-language models in explainable biomedical AI, a topic largely underexplored in previous work. Our contributions also include a summary of widely used evaluation metrics and open-source frameworks, along with a critical discussion of persistent challenges and future directions. This survey offers a timely and in-depth foundation for advancing interpretable DL in biomedical image analysis.
\end{abstract}

\ccsdesc[500]{Applied computing~Health informatics}
\ccsdesc[300]{Computing methodologies~Artificial intelligence}

\keywords{Explainable AI (XAI), Interpretability, Deep Learning, Biomedical Image Analysis}

\maketitle

\section{Introduction}
\label{sec:introduction}
Biomedical image analysis is fundamental to modern medicine and life science research, enabling the extraction of critical information from diverse imaging modalities \cite{Singh2020}.
These include conventional medical images such as X-ray, CT, MRI, and ultrasound, as well as specialized biological images like histopathological slides, fluorescence microscopy images, and cellular or genomic visualizations. Recent advances in computational hardware and the availability of large-scale biomedical imaging datasets have driven the rapid adoption of deep learning (DL) techniques in this field \cite{suganyadevi2022review}. DL models have shown exceptional success in tasks such as tumor detection, organ segmentation, lesion classification, and cellular-level disease characterization \cite{zhu2022compound,zhu2024ac}.
By capturing complex patterns and subtle variations often imperceptible to human observers, these models significantly enhance diagnostic accuracy, processing efficiency, and scalability \cite{zhu2020neural, Young2022}.
As biomedical research and clinical applications continue to evolve, DL has become an indispensable tool for enhancing precision medicine and advancing scientific discovery.

Despite the success, DL models' inherent opacity remains a significant obstacle to clinical integration \cite{Jia2019}. The so-called ``black-box” nature of DL raises concerns among clinicians, researchers, and regulatory bodies regarding model transparency and accountability \cite{Dhar2023}. In healthcare, decision-making must be explainable and justifiable. Clinicians are not only expected to interpret algorithmic outputs but also to communicate and defend their decisions to patients. When the reasoning behind a model’s predictions is inaccessible, trust in its outputs diminishes, especially in scenarios where errors may lead to serious or irreversible consequences. This interpretability gap undermines both user confidence and patient safety, thereby limiting the broader adoption of DL models in clinical and biomedical practice.

\begin{figure}[htbp]
  \centering
  \vspace{-0.1cm}
  \includegraphics[width=0.56\textwidth]{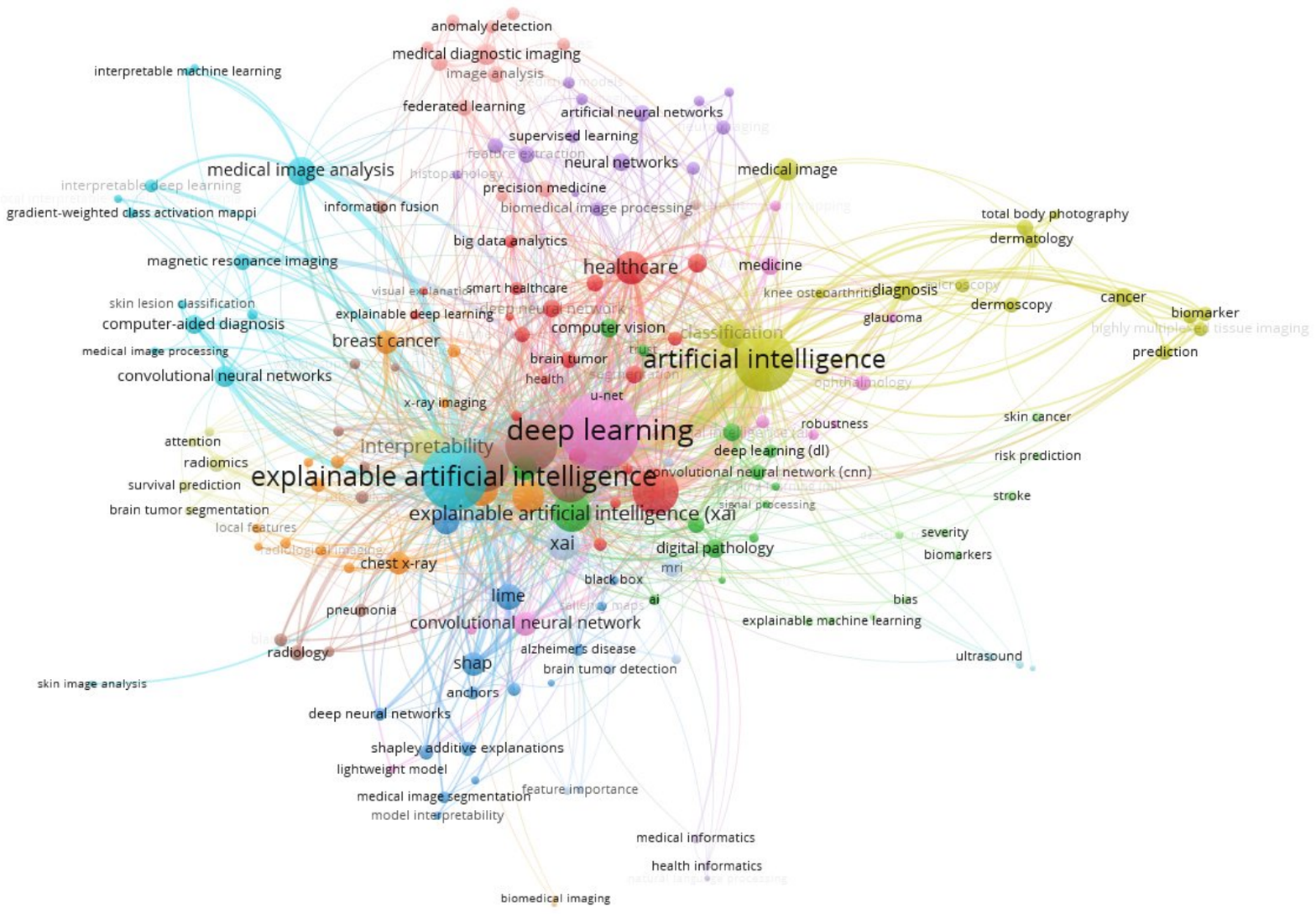}
  \vspace{-0.3cm}
  \caption{Keyword co-occurrence network for XAI in biomedical image analysis, generated using \textsc{VOSviewer}. Node size and edge density reflect term relevance and centrality. The prominence of DL and XAI highlights the increasing integration of XAI in this domain.}
  \vspace{-0.4cm}
  \label{fig:xai_gain_network}
\end{figure}

To address the challenge of model opacity, explainable artificial intelligence (XAI) has emerged as a promising direction \cite{Ayesha2021}. XAI techniques aim to enhance the transparency of DL models by elucidating their internal reasoning processes and generating human-understandable explanations for model predictions \cite{Sheu2022}.
In the context of biomedical image analysis, where trust, accountability, and precision are critical, XAI has gained increasing attention (Fig.~\ref{fig:xai_gain_network}). These methods are being applied across various biomedical imaging tasks and modalities, supporting not only interpretability but also model validation, bias detection, and regulatory compliance. However, despite the growing body of literature, current research remains fragmented across different application domains and methodological frameworks. There is a clear need for a comprehensive survey that systematically consolidates current developments, categorizes existing XAI methods, and outlines challenges and opportunities specific to biomedical image analysis. This paper addresses this gap by providing a structured and in-depth review of the field, with the goal of guiding future research and promoting the safe and transparent deployment of DL models in biomedical image analysis.

\subsection{Related Surveys}
Several prior surveys have explored XAI in medical image analysis, but most exhibit notable limitations. The review by \cite{Singh2020} was published before the surge of post-2020 developments and is now outdated. \cite{Salahuddin2022} lacks modality-specific analysis and omits emerging topics such as foundation models and open-source frameworks, which are increasingly central to biomedical image analysis. \cite{Borys12023} only focuses on vision-based XAI methods, while \cite{Borys2023} covers only non-vision-based methods, and neither provides an in-depth discussion of XAI across diverse biomedical tasks or detailed methodological insights. \cite{patricio2023explainable} includes both vision- and non-vision-based methods but limits its scope to only classification and overlooks key tasks such as segmentation, detection, and the interpretation of pathology or cellular microscopy images. It also lacks technical depth, recent advances in foundation models, and comparisons across different imaging modalities.

The closest works to ours are \cite{vanderVelden2022} and \cite{Nazir2023}. However, \cite{vanderVelden2022} organizes XAI methods by anatomical location, lacking a thorough methodological review and modality-specific interpretability requirement analysis. This anatomy-driven perspective overlooks the methodological differences between image modalities. For example, in X-ray images, interpretation typically focuses on localized density changes, whereas in ultrasound images, due to significant noise and variable image quality, XAI must address greater uncertainties. As a result, this review lacks both the breadth and methodological depth. Although \cite{Nazir2023} discusses XAI methods from the perspective of image modality, it lacks a systemic review of XAI techniques, provides limited linkage between XAI methods and specific biomedical imaging tasks, omits recent advancements in foundation models, and covers only a narrow range of image types. 

To address these gaps, this survey presents a comprehensive and up-to-date review of XAI in biomedical image analysis. We systematically categorize and compare existing XAI methods based on their underlying principles, advantages, and limitations, and highlight recent advances and emerging trends in the field. Building on this foundation, we introduce a novel modality-centered taxonomy that aligns XAI techniques with specific imaging modalities, highlighting the distinct interpretability challenges across imaging modalities. We further extend our review to recent progress in multimodal learning and vision-language models (VLMs), which are growing important yet remain underexplored in prior surveys. To support practical adoption, we also summarize commonly used evaluation metrics and open-source XAI frameworks for biomedical applications. By offering both technical depth and practical insights, this survey fills critical gaps in the literature and establishes a foundation for advancing interpretable DL in biomedical image analysis. A  comparative summary of existing surveys and our contributions is presented in Table~\ref{tab:xai_survey_comparison}.

\begin{table}[htbp]
\caption{Comparison of existing surveys and this work.}
\vspace{-0.3cm}
\label{tab:xai_survey_comparison}
\renewcommand{\arraystretch}{0.9}
\resizebox{0.84\textwidth}{!}{
\begin{tabular}{llcccccccc}
\toprule[1pt]
\multicolumn{2}{l}{\multirow{2}{*}{\textbf{Comparison Criteria}}} & 
\multicolumn{7}{c}{\textbf{Related Surveys}} & 
\multirow{2}{*}{\textbf{Ours}} \\
\cline{3-9}
& & \cite{Singh2020} & 
\cite{vanderVelden2022} & \cite{Salahuddin2022}  & \cite{Borys2023}  & \cite{Borys12023}  & \cite{patricio2023explainable} & \cite{Nazir2023} &  \\
\midrule[1pt]
\multirow{4}{*}{Visualization-based XAI} 
& CAM-based                     & \pmark & \pmark & \pmark & \xmark & \cmark & \pmark & \pmark & \cmark  \\
& Grad/Backpropagation-based   & \pmark & \cmark & \cmark & \xmark & \cmark & \pmark & \cmark & \cmark  \\
& Attention-based              & \pmark & \xmark & \pmark & \xmark & \xmark & \pmark & \pmark & \cmark  \\
& Perturbation-based           & \pmark & \cmark & \cmark & \xmark & \cmark & \pmark & \cmark & \cmark  \\
\hline
\multirow{3}{*}{Non-visualization-based XAI}
& Example-based                & \xmark & \cmark & \pmark & \cmark & \xmark & \cmark & \cmark & \cmark  \\
& Concept-based                & \pmark & \pmark & \cmark & \cmark & \xmark & \cmark & \cmark & \cmark  \\
& Text-based                   & \pmark & \cmark & \pmark & \cmark & \xmark & \cmark & \cmark & \cmark  \\
\hline
\multicolumn{2}{l}{Latent-based XAI}   & \xmark & \xmark & \pmark & \cmark & \xmark & \pmark & \pmark & \cmark  \\
\hline
\multirow{8}{*}{Modality-Specific }
& Radiographic           & \pmark & \cmark & \xmark & \xmark & \xmark & \pmark & \cmark & \cmark  \\
& CT                           & \pmark & \cmark & \xmark & \xmark & \xmark & \pmark & \cmark & \cmark  \\
& MRI                          & \pmark & \cmark & \xmark & \xmark & \xmark & \xmark & \cmark & \cmark  \\
& Ultrasound                   & \xmark & \cmark & \xmark & \xmark & \xmark & \cmark & \cmark & \cmark  \\
& PET                          & \xmark & \xmark & \xmark & \xmark & \xmark & \xmark & \pmark & \cmark \\
& Optical      & \pmark & \cmark & \xmark & \xmark & \xmark & \pmark & \cmark & \cmark  \\
& Microscopy        & \xmark & \pmark & \xmark & \xmark & \xmark & \pmark & \pmark & \cmark  \\
& Multi-Modality              & \xmark & \xmark & \xmark & \xmark & \xmark & \xmark & \xmark & \cmark \\
\hline
\multicolumn{2}{l}{Interpretable Vision-Language Models}   & \xmark & \xmark & \xmark & \xmark & \xmark & \xmark & \xmark  &\cmark \\ \hline
\multicolumn{2}{l}{Open-source Frameworks}     & \xmark & \xmark & \xmark & \pmark & \pmark & \cmark & \pmark  & \cmark \\ \hline
Evaluation Criteria           &   & \xmark & \pmark & \pmark & \xmark & \xmark & \cmark & \cmark  & \cmark \\
\bottomrule[1pt]
\end{tabular}
}
\caption*{\textbf{Note:} \cmark = covered; \pmark = partially covered (e.g., briefly mentioned or lacking full discussion); \xmark = not mentioned.}
\vspace{-0.7cm}
\end{table}

\subsection{Contribution}
The main contributions of this survey are summarized as follows:
\begin{itemize}
\item We provide a systematic classification and in-depth analysis of existing XAI techniques, specifically tailored to biomedical image analysis. By examining their methodological foundations, advantages, and limitations within biomedical contexts, we offer a structured technical landscape to support informed method selection.

\item Unlike prior surveys, we propose a modality-centered taxonomy that maps XAI methods to specific biomedical imaging modalities, revealing the distinct interpretability challenges and requirements of each. This modality-aware perspective enables targeted application of XAI techniques to diverse biomedical tasks.

\item We extend the scope of traditional XAI reviews by incorporating recent advances in multimodal learning and VLMs, two rapidly evolving research directions in biomedical AI. This timely discussion anticipates future trends and highlights the increasing complexity of explainability in data-rich biomedical environments.

\item We create a curated repository of open-source XAI frameworks and summarize widely used evaluation metrics for interpretability. These resources support reproducibility and adoption, enable consistent benchmarking, and facilitate the development and deployment of explainable models in biomedical applications.

\item We identify and analyze persistent challenges unique to XAI in biomedical image analysis, issues often overlooked in existing reviews. Building on this critical perspective, we outline open research directions to advance the development of interpretable and domain-aligned DL systems.
\end{itemize}

\subsection{Literature Collection and Selection}
To ensure comprehensiveness and scientific rigor, we adopted a structured search strategy to identify peer-reviewed publications on the integration of XAI methods in DL-based biomedical image analysis. Literature was retrieved from major databases, \textit{Scopus}, \textit{PubMed}, \textit{Google Scholar}, \textit{IEEE Xplore}, and \textit{Web of Science}, using carefully formulated Boolean queries such as ``\textit{(explainable OR interpretable) AND (AI OR deep learning) AND medical AND image}". To ensure modality coverage, additional keywords related to specific imaging types, such as radiological and microscopic imaging, were incorporated. Eligible studies focused on the application or development of XAI methods for biomedical image analysis tasks. Articles unrelated to imaging or lacking a substantial discussion of explainability were excluded.

\subsection{Structure of the Paper}
The remainder of this survey is organized as follows. Section~\ref{sec2} classifies existing XAI methods, examining their foundations, strengths, and limitations. Section~\ref{sec3} introduces a novel modality-centered taxonomy linking XAI techniques to specific biomedical imaging types, and extends the discussion to emerging directions in multimodal learning and vision-language models. Section~\ref{sec4} presents a curated collection of open-source XAI frameworks, while Section~\ref{sec5} reviews commonly used interpretability evaluation metrics to support reproducibility and benchmarking. Section~\ref{sec6} outlines open challenges and future directions in the biomedical imaging context, followed by concluding insights in Section~\ref{sec7}.

\section{Taxonomy of XAI Methods}\label{sec2}
\begin{figure}[htbp]
  \centering
  \vspace{-0.2cm}
  \includegraphics[width=0.65\textwidth]{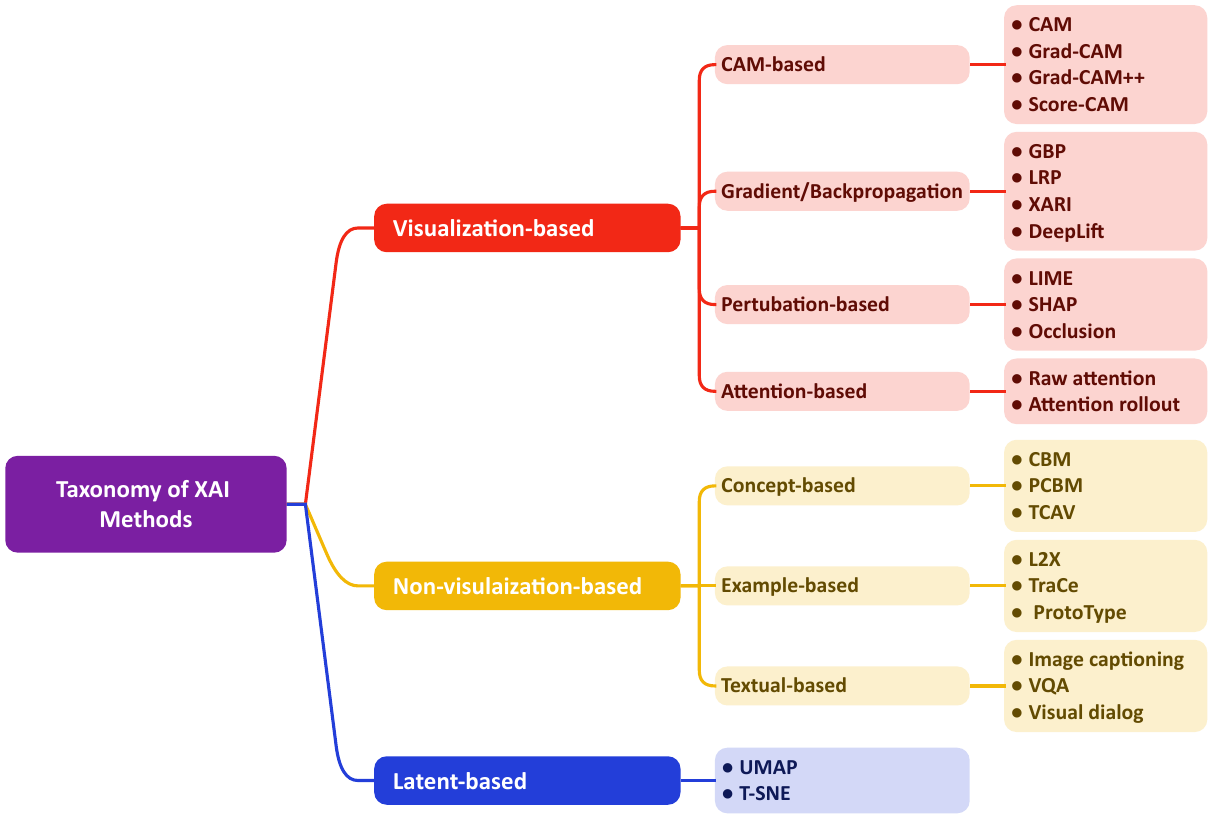}
  \vspace{-0.2cm}
  \caption{A structured taxonomy of XAI methods in biomedical image analysis.}
  \vspace{-0.4cm}
  \label{fig:xai_taxonomy}
\end{figure}

XAI was first introduced by \cite{van2004explainable} and has evolved into a broad set of techniques for enhancing the transparency and interpretability of DL models. XAI methods are typically categorized along several dimensions, such as intrinsic or post-hoc, global or local, and model-specific or model-agnostic. In biomedical image analysis, visual or semantic justifications are often required to support medical decision-making. To structure the landscape of XAI in this domain, we propose a taxonomy comprising three major categories (Fig.~\ref{fig:xai_taxonomy}): visualization-based methods, which highlight spatial features to provide intuitive visual cues; non-visualization-based methods, which rely on example-based reasoning, concept-level abstraction, or natural language generation; and latent-based methods, which explain model behavior through analysis of latent feature representations. This taxonomy reflects the diverse interpretability demands in biomedical imaging, where the balance between visual clarity and semantic depth is essential for clinical utility.

\subsection{Visualization-Based Methods}\label{subsec1}
Visualization-based XAI methods explain model predictions by highlighting spatial regions that influence decision-making, typically via heatmaps or saliency maps (Fig.~\ref{fig:grad_cam_pipeline}). These intuitive visualizations are especially valuable in biomedical image analysis, where clinical visual interpretability is essential. We categorize these methods into four groups: class activation mapping (CAM)-based, gradient-based, perturbation-based, and attention-based. 

\subsubsection{CAM-Based Methods}
\label{subsubsec:CAM}
\textbf {CAM} is a foundational method for visualizing the image regions most influential to a DL model’s decision. Introduced by \cite{7780688}, CAM generates class-specific heatmaps by computing a class score-weighted summation of the feature maps from the final convolutional layer. Specifically, given the activation $A_k(x, y)$ of the $k$-th feature map at position $(x, y)$ and its associated weight $w_k^{c}$ for class $c$, the heatmap $L^{c}_{CAM}(x, y)$ is computed as: 
\begin{math}
     L^{c}_\text{CAM}(x, y) = \sum\limits_k w_k^{c} A_k(x, y).
\end{math}
Despite its interpretability, CAM is limited to architectures with global average pooling (GAP) and fully connected layers, typically requiring model retraining to incorporate these architectural constraints.

\begin{figure*}[tbp]   
  \centering
  \includegraphics[width=0.85\textwidth]{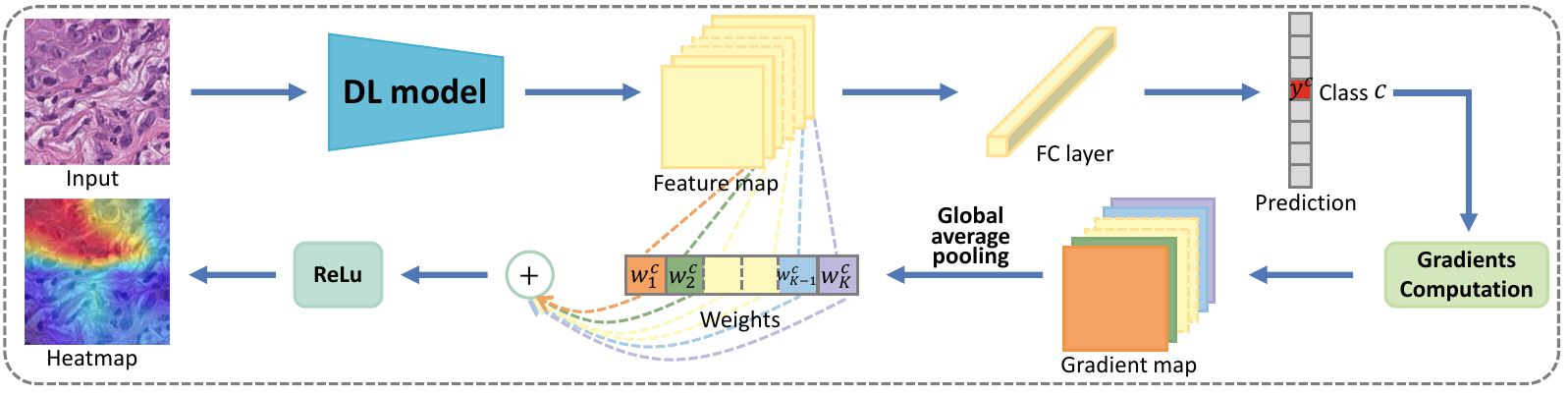}
  \vspace{-0.3cm}
  \caption{Grad-CAM workflow for generating class-specific visual explanations. Gradients of the target class score with respect to convolutional feature maps are globally averaged to compute importance weights, which are combined with feature maps and passed through ReLU to produce heatmaps highlighting discriminative regions.}
  \label{fig:grad_cam_pipeline}
\end{figure*}

\textbf{Gradient-weighted CAM (Grad-CAM)} addresses the architectural limitations of CAM by removing the dependency on GAP, thus making it applicable to to a wider range of network architectures \cite{8237336}. As illustrated in Fig.~\ref{fig:grad_cam_pipeline}, it computes the gradient of the target class score $y^{c}$ with respect to the feature maps $A_k$ of a selected convolutional layer. The importance weight $w_k^{c}$ for feature map $A_k$ is obtained by globally averaging the gradients:
\begin{math}
    w_k^{c} = \frac{1}{H \cdot W} \sum\limits_{x=1}^H \sum\limits_{y=1}^W
        \frac{\partial y^{c}}{\partial A_k(x, y)},
\end{math}
where $H$ and $W$ are the feature map sizes. The final heatmap is computed similarly to CAM, followed by a ReLU operation to retain only positive contributions: \begin{math}
    L^{c}_\text{Grad-CAM}(x, y) = \text{ReLU}\Big(\sum\limits_k w_k^{c} A_k(x, y)\Big).
\end{math}
While Grad-CAM enhances flexibility and generalization across architectures, it may face limitations in multi-task or multi-label settings, where overlapping gradients computed on shared feature maps can lead to ambiguous attribution and reduced class interpretability.

\textbf{Grad-CAM++} extends Grad-CAM to enhance spatial precision, especially in cases involving multiple objects or fine-grained features \cite{8354201}. It refines the gradient weighting mechanism using higher-order derivatives, allowing the heatmaps to more accurately capture fine-grained contributions of individual pixels to model prediction. The class-specific heatmap retains the same form as in CAM, but the importance weight $w_k^{c}$ is refined as:
\begin{math}
        w_k^{c} = \sum\limits_{x=1}^H \sum\limits_{y=1}^W \alpha_k^{c}(x, y) \cdot
        \text{ReLU}\Big(\frac{\partial y^{c}}{\partial A_k(x, y)}\Big),
\end{math}
where $\alpha_k^{c}(x, y)$ is a data-dependent weighting factor computed using second- and third-order partial derivatives.
While Grad-CAM++ improves spatial precision and class discrimination, the use of higher-order gradients may introduce numerical instability, particularly when applied to deep architectures or large-scale datasets.

\textbf{Smooth Grad-CAM++} extends Grad-CAM++ by enhancing the stability and visual coherence of class activation maps using stochastic smoothing \cite{omeiza2019smooth}. Specifically, it applies Gaussian noise to the input multiple times, computes Grad-CAM++ heatmaps for each perturbed sample, and averages the resulting heatmaps. This noise-averaging strategy suppresses high-frequency artifacts and yields smoother, more stable explanations, which are particularly beneficial in fine-grained biomedical imaging tasks. Despite its improved smoothness and noise robustness, it introduces considerable computational overhead and requires careful tuning of hyperparameters such as noise magnitude and sample count.

\textbf{Axiom-Based Grad-CAM (XGrad-CAM)} builds upon Grad-CAM by introducing a set of interpretability axioms, such as sensitivity, implementation invariance, and graduality, to improve the consistency and human-alignment of visual explanations \cite{fu2020axiombased}. Rather than relying solely on raw gradient values, it modifies the importance weights by explicitly integrating feature activations, thereby emphasizing contributions that are both strong and semantically meaningful. The importance weight $w_k^{c}$ is redefined as:
\begin{math}
        w_k^{c} = \sum\limits_{x=1}^H \sum\limits_{y=1}^W
        \Big( \frac{\partial y^{c}}{\partial A_k(x, y)} \cdot
        \frac{A_k(x, y)}{\sum\limits_{x=1}^H \sum\limits_{y=1}^W A_k(x, y)} \Big).
\end{math}
By weighting gradients based on normalized activation values, XGrad-CAM aims to more faithfully reflect the spatial importance of each feature map. However, its effectiveness may vary depending on network architecture and task characteristics, and the selection or tuning of axioms often requires domain expertise and empirical validation.

\textbf{Score-Weighted CAM (Score-CAM)} is a gradient-free extension of CAM that enhances both heatmap accuracy and interpretability by using class scores as importance weights \cite{9150840}. Instead of  relying on gradients, it directly evaluates each activation map's contribution to model's prediction by measuring the change in class confidence when the input is masked by that map. Its importance weight $w_k^{c}$ is computed as: \begin{math}
     w_k^{c} = softmax\Big(y^{c}(M_k) - y^{c}(X_b)\Big)_k,
\end{math}
where $y^{c}(M_k)$ and $y^{c}(X_b)$ are the predicted scores for class $c$ with masked input $M_k$ and baseline input $X_b$, respectively. The masked input $M_k$ is given by: \begin{math}
   M_k = \frac{U(A_k) - \min\limits U(A_k)}{\max\limits  U(A_k) - \min\limits  U(A_k)}
        \odot X_b, 
\end{math}
where $U$ is the upsampling and $\odot$ element-wise multiplication. While Score-CAM improves robustness and is model-agnostic, especially when gradients are unreliable, it is computationally intensive and sensitive to baseline selection and softmax scaling, which may affect consistency across settings.

\textbf{Smoothed Score-CAM (SS-CAM)} improves Score-CAM by introducing input perturbation to improve attribution stability and reduce sensitivity to local variations \cite{wang2020sscam}. Specifically, it injects Gaussian noise into the input image multiple times and averaging the resulting class scores across $N$ perturbations, thereby generating smoother and more robust importance weights. The weight $w_k^{c}$ is computed as:
\begin{math}
     w_k^{c} = softmax\Big(\frac{1}{N} \sum\limits_{i=1}^N (y^{c}(\hat{M_k}) - y^{c}(X_b))\Big)_k, 
\end{math}
where the noise-augmented masked input $\hat{M_k}$ is computed by:
\begin{math}
    \hat{M_k} = (\frac{U(A_k) - \min\limits U(A_k)}{\max\limits  U(A_k) - \min\limits  U(A_k)} + \delta) \odot X_b,
\end{math}
with $\delta \sim \mathcal{N}(0, \sigma^2)$ denoting the Gaussian noise. Although SS-CAM improves smoothness and stability, its performance depends on task complexity and network architecture, and it incurs additional costs from repeated forward passes.
   
\textbf{Integrated Score-CAM (IS-CAM)} enhances Score-CAM by integrating integrated gradients to capture more complete and stable feature attributions across a range of input perturbations \cite{naidu2020iscam}. Instead of relying on a single perturbed input, IS-CAM applies a series of perturbations and aggregates the resulting class score differences to compute importance weights. Its weight $w_k^{c}$ is computed by the same form as SS-CAM, but each perturbed masked input $\hat{M_k}$ is constructed via: 
\begin{math}
    \hat{M_k} = \sum\limits_{j=0}^{i-1} \frac{j}{N}
        \frac{U(A_k) - \min\limits U(A_k)}{\max\limits  U(A_k) - \min\limits  U(A_k)} \odot X_b.
\end{math}
By integrating across multiple input states, IS-CAM enhances attribution completeness but incurs high computational costs due to the large number of forward passes required for each perturbation step, limiting its practicality in real-time or resource-constrained settings.

\textbf{Layer-CAM} advances the CAM family by generating fine-grained, spatially aware heatmaps through pixel-level weighting across convolutional layers \cite{9462463}. Unlike prior CAM variants that rely on globally averaged gradients from the final convolutional layer, Layer-CAM computes location-specific importance scores using intermediate activations, thereby capturing hierarchical and localized model responses. The class-specific heatmap is computed as: \begin{math}
    L^{c}_\text{Layer-CAM}(x, y) = \text{ReLU}\Big(\sum\limits_k w_k^{c}(x, y) \cdot A_k(x, y)\Big),
\end{math}
where the spatially varying importance weight $w_k^{c}(x, y)$ is defined as: \begin{math}
     w_k^{c}(x, y) = \text{ReLU}\Big(\frac{\partial y^{c}}{\partial A_k(x, y)}(x, y)\Big).
\end{math}
This pixel-level attention enables more precise localization and enhances interpretability, especially when applied across multiple intermediate layers. However, its effectiveness depends on the choice of layers and incurs higher computational cost due to the need for layer-wise gradient backpropagation.

\textbf{Ablation-CAM} is a gradient-free CAM variant that generates heatmaps by ablating feature maps to assess their contribution to the model’s prediction \cite{9093360}. Instead of relying on backpropagated gradients, it computes class-specific weights by quantifying the drop in prediction confidence when a given feature map is removed, thereby reducing the noise and instability commonly associated with gradient-based methods. The weight is calculated as:
$ w^{c}_k =\frac{y^{c}-y^{c}_{k}}{y^{c}}, $
where $y^{c}_{k}$ is the score for class $c$ without the $k$-th feature map. By quantifying the output degradation from feature suppression, Ablation-CAM yields more stable and interpretable explanations, especially useful for intra-class attribution. However, it is computationally expensive, as it requires multiple forward passes, one for each ablated feature map.

\textbf{Eigen-CAM} introduces a gradient-free approach to CAM by leveraging principal component analysis (PCA) on feature maps \cite{Muhammad_2020}. Instead of relying on class-specific gradients or ablation, Eigen-CAM identifies the most dominant patterns in feature representations through PCA, enabling the generation of class-agnostic yet highly informative heatmaps. The heatmap is computed as:
$L_{\mathrm{Eigen-CAM}(x,y)} = \sum_k P_k \cdot A_k(x,y), $
where $P_k$ is the weight of the $k$-th feature map derived from the first principal component of the feature map matrix. By projecting high-dimensional activations onto their principal direction, Eigen-CAM suppresses noise and emphasizes salient structures, enabling efficient visualization. However, the PCA-based projection may lead to information loss and increased computational overhead, particularly when applied to large-scale or multi-layer networks.

\subsubsection{Gradient/Backpropagation-Based Methods}
This category interprets DL models by analyzing output gradients with respect to internal activations. Unlike CAM-based methods, which use gradients for spatial localization, these techniques use gradients to attribute predictions at the feature level, offering deeper insight into feature importance.

\textbf{DeconvNet} is a gradient-based method that projects feature activations back to the input space to reveal input patterns that activate specific neurons \cite{zeiler2014visualizing}. Through a sequence of unpooling, rectification, and deconvolution, it visualizes the hierarchical features learned by convolutional neural networks (CNNs). However, it is mainly applicable to CNNs and does not generalize well to other network types.

\textbf{Guided Backpropagation (GBP)} is a refinement of DeconvNet that modifies the standard backpropagation process to produce sharper and more interpretable saliency maps \cite{springenberg2014striving}. The key idea is to guide the backward flow of gradients by suppressing negative gradients at ReLU layers, thereby highlighting input features that positively contribute to the model’s prediction. While GBP enhances visual clarity, it is sensitive to noise, introduces additional computational complexity, and is less effective for non-convolutional architectures such as recurrent neural networks.

\textbf{Layer-wise Relevance Propagation (LRP)} decomposes a model’s prediction by attributing relevance scores to individual input features \cite{Bach2015}. It systematically propagates the prediction score backward through the network, layer by layer, redistributing the relevance of each neuron to its predecessors based on their contribution to the activation. For a neuron $j$ in layer ${l+1}$, the relevance $R_{i}^{(l+1)}$ is redistributed to its input neurons $i$ in previous layer $l$ according to: 
$ R_{i}^{(l)} = \sum_{j}^{ } \frac{x_{i}w_{ij}}{\sum_{i}^{}x_{i}. w_{ij}+\epsilon\cdot \text{sign}(\sum_i x_i w_ij)}R_{j}^{(l+1)},$
where $x_i$ is the activation of neuron $i$, $w_{ij}$ is the weight connecting neuron $i$ to $j$, $\epsilon$ is a small constant to improve numerical stability, and sign($\cdot$) is the sign function used to maintain the sign of the values. The use of the $\epsilon$-rule helps ensure the robustness of relevance propagation, especially in deep architectures. However, the effectiveness of LRP can vary depending on the network structure and hyperparameter configurations.

\textbf{Integrated Gradients (IG)} is designed to address limitations of standard gradient techniques, such as gradient saturation and noise sensitivity \cite{sundararajan2017axiomatic}. IG attributes the model prediction to input features by integrating the gradients along a straight-line path between a baseline input $\mathbf{x'}$ and the actual input $\mathbf{x}$. The baseline is typically a zero vector or another neutral reference. The attribution for input feature $x_i$ is computed as:
  $  \text{IG}_{i}(\mathbf{x}) =(x_{i}-x^{'}_{i})\int_{\alpha =0}^{1}\frac{\partial F(\mathbf{x'}+\alpha(\mathbf{x}-\mathbf{x'}))}{\partial x_{i}}d\alpha,$
where $IG_{i}(\mathbf{x})$ is the integrated gradient along the integral path $\mathbf{x'}+\alpha(\mathbf{x}-\mathbf{x'})$, $F$ is the model's prediction function, and $\alpha$ is distributed in range $[0,1]$. 
IG provides smooth and axiomatic feature attributions, but it does not capture feature interactions and is sensitive to the choice of the baseline input, which can affect interpretability in practice.
 
\textbf{Explainable Representations through AI (XRAI)} is a region-based extension of IG designed to generate more semantically meaningful saliency maps \cite{kapishnikov2019xrai}. Unlike pixel-level attribution methods, XRAI segments the input image into interpretable regions and attributes relevance scores to these regions based on IG-derived gradients. This region-level aggregation aligns better with human perceptual understanding and improves interpretability in visual tasks. However, its reliance on repeated IG computations across multiple image segments leads to high computational complexity, making it less efficient for large-scale models or high-resolution inputs.

\textbf{Deep Learning Important FeaTures (DeepLIFT)} explains DL model predictions by propagating activation differences between the input and a reference baseline \cite{shrikumar2017learning}. Unlike standard gradients that compute local sensitivity, DeepLIFT attributes prediction based on the difference in activation relative to the baseline, thereby addressing issues such as gradient saturation. For input $x_i$ and baseline $x'_i$, the difference $\Delta x_i = x_i -x'_i$ is use to compute contribution scores $C_{\Delta x_i \Delta t}$, such that the output difference $\Delta t = \sum_i C_{\Delta x_i \Delta t}$, where $C_{\Delta x_i \Delta t} = \frac{\partial t}{\partial x_i}\Delta x_i$. Here, $\frac{\partial t}{\partial x_i}$ approximates the gradient of the output $t$ with respect to $x_i$. While DeepLIFT improves attribution reliability over traditional gradients, its interpretability is sensitive to the choice of reference input, which can substantially influence the resulting explanations.

\subsubsection{Attention-Based Methods}
With the rise of Transformer architectures in medical image analysis, attention-based XAI methods have gained prominence by leveraging Transformers' inherent self-attention to produce intrinsically interpretable, model-specific explanations. These methods visualize attention weights or quantify attention flow to reveal how information propagates across layers. Abnar et al. \cite{abnar-zuidema-2020-quantifying} introduced Attention Rollout and Attention Flow, which trace input contribution through cumulative attention, offering more faithful feature attributions by simulating information propagation from input to output. However, due to their simplifying assumptions and cumulative effects, they suffer from high computational costs and risk attributing relevance to irrelevant input regions. To address this, \citet{playout2022focused} introduced Focused Attention, a method that generates high-resolution attribution maps through iterative conditional patch resampling. This approach selectively amplifies the most informative image regions, improving the spatial precision and interpretability of attention-based explanations. Despite these advances, attention weights may not reliably reflect model reasoning, particularly in deeper layers where attention can become diffuse or uniform.

\subsubsection{Perturbation-Based Methods} 
These methods interpret DL models by modifying inputs and observing changes in predictions. Without relying on model gradients or architecture, they estimate feature importance based on output sensitivity to localized or structured input perturbations. These methods offer intuitive, model-agnostic explanations and are broadly applicable across architectures. 

\textbf{Local Interpretable Model-Agnostic Explanations (LIME)} explains individual predictions by approximating a complex model's local behavior with an interpretable surrogate, typically a linear regressor \cite{ribeiro2016should}. It generates perturbed samples around a given input, obtains their predictions from the original model, and then fits the surrogate to estimate the importance of each input. While LIME offers intuitive, input-specific explanations, it is inherently local and may fail to capture a model’s global decision boundaries. Furthermore, it is sensitive to the sampling strategy and the fidelity of the surrogate model, which can lead to variability and potential instability in the generated explanations.

\textbf{Shapley Additive Explanation (SHAP)} is an attribution method grounded in cooperative game theory \cite{lundberg2017unified}. It evaluates the importance of each input feature to a model's prediction by computing its Shapley values, which represents the average marginal contribution of the feature across all possible feature subsets. It provides theoretically sound and locally accurate explanations, but exact computation of Shapley values is computationally expensive for high-dimensional inputs. To address this, DeepSHAP \cite{chen2020explaining} combines SHAP principles with backpropagation-based heuristics, offering a more efficient approximation for deep models.

\textbf{Anchor} is a rule-based explanation method that generate high-precision if-then rules, called anchors, to identify input feature subsets which, when fixed, lead to consistent model predictions with high probability, even when other parts of the input are perturbed \cite{ribeiro2018anchors}. These human-readable rules provide intuitive explanations of model behavior. It is particularly effective for models with clear decision boundaries but may struggle with complex, highly nonlinear models where local consistency is harder to maintain. Additionally, the process of searching for and validating anchor rules is computationally intensive, especially when applied to high-dimensional or large-scale datasets.

\textbf{Randomized Input Sampling for Explanation (RISE)} explains model predictions by randomly occluding different regions of the input image and then passing each masked image through the model to observe changes in the output to infer which input regions are most important to the model's decision \cite{petsiuk2018rise}. It does not rely on gradients or internal model information, making it applicable to any black-box model. However, it incurs significant computational cost due to the large number of forward passes required. Moreover, its reliance on coarse, random masks may limit its ability to capture fine-grained feature importance, especially in high-resolution images or subtle decision boundaries.

\textbf{Similarity Difference and Uniqueness (SIDU)} evaluates input feature importance by jointly measuring similarity difference and uniqueness \cite{muddamsetty2020sidu}. It perturbs input images using spatial masks derived from the last convolutional layer and observes the impact of these perturbations on model’s prediction confidence. Similarity difference quantifies how much a region influences the output compared to the original input, while uniqueness quantifies how distinct its influence is relative to other regions. By combining these two metrics, it produces fine-grained saliency maps that highlight both discriminative and distinctive regions. However, SIDU is computationally intensive due to extensive perturbations and forward passes, limiting scalability to high-resolution data or large-scale datasets.

\textbf{Occlusion} assesses feature importance by systematically masking regions of the input and measuring the resulting change in model predictions \cite{zeiler2014visualizing}. By occluding one region at a time, it identifies areas critical to model prediction. This method requires no access to model internals, making it broadly applicable across architectures. However, its computational cost grows with image resolution, as each occluded region necessitates a separate forward pass, hindering its practicality for high-resolution images or large-scale datasets.

\textbf{Meaningful Perturbation} evaluates feature importance by learning an optimal perturbation mask that identifies regions most influential to the model’s prediction, offering a more principled alternative to random or constant-value masking.
Unlike brute-force occlusion, it introduces semantically meaningful changes to the input. However, in medical imaging, applying such perturbations is problematic, as substituting regions with unrealistic patterns (e.g., constant values) can lead to artifacts that distort clinical relevance.
\citet{dabkowski2017real} introduced a real-time variant that approximates the perturbation mask through a single forward pass to find the ideal perturbation mask.

\subsection{Non-Visualization-Based Methods}
\label{subsec2}
\begin{figure*}[htbp]
  \centering
  \vspace{-0.1cm}
  \includegraphics[width=0.8\textwidth]{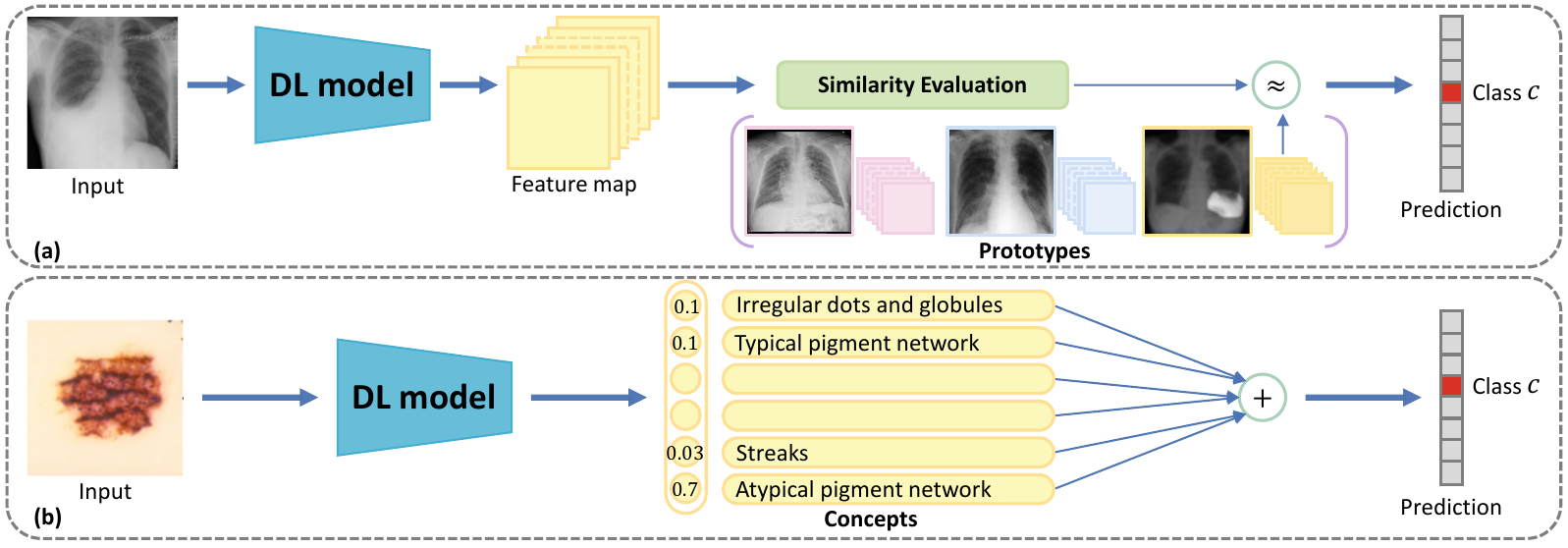}
  \vspace{-0.3cm}
  \caption{Illustration of two non-visual XAI paradigms in medical image analysis. (a) Example-based XAI: The input is compared in latent space to a set of learned prototype examples, with prediction informed by the most similar prototypes. (b) Concept-based XAI: Deep features are mapped to a space of clinically meaningful concepts, whose activations contribute to the final prediction. }
  \vspace{-0.3cm}
  \label{fig:concept}
\end{figure*}

Non-visual XAI methods explain DL model predictions using representative examples, high-level concepts, or natural language, rather than spatial saliency maps (Fig.~\ref{fig:concept}). We categorize them into three groups: example-based, concept-based, and text-based.

\subsubsection{Example-Based Methods}
\textbf{Conceptual and Counterfactual Explanations (CoCoX)} combines concept-based and counterfactual reasoning to interpret model decisions \cite{Akula2020}. It identifies ``fault-lines", defined as the minimum set of semantic features whose addition or removal alters the model’s prediction. These are categorized as positive or negative fault-lines, indicating supportive or opposing influences on the current classification. By tracing these concept-level changes, CoCoX offers semantically grounded explanations for why a model predicts a particular class. However, its fidelity depends on the quality and completeness of learned concept representations.

\textbf{Counterfactual Explanations Guided by Prototypes} generates counterfactual explanations \cite{van2021interpretable}. It identifies prototypical instances from the training data to guide the generation of semantically meaningful counterfactual examples, i.e., minimal input modifications that would lead to a change in the model’s prediction. These prototypes serve as anchors to ensure the counterfactuals are both realistic and interpretable. However, the method’s effectiveness is contingent on the representativeness and diversity of the prototype set; poorly distributed or uninformative prototypes can undermine the reliability and interpretability of the resulting explanations.

\textbf{Contrastive Explanations Method (CEM)} interprets model predictions by identifying which features are essential or irrelevant for a given prediction \cite{dhurandhar2018explanations}. Specifically, it identifies ``Pertinent Positives", features minimally sufficient for the current decision, and ``Pertinent Negatives", features whose absence preserves the prediction. This contrastive approach clarifies both why a decision was made and why alternatives were rejected. While CEM provides intuitive, counterfactual-style insights, it is computationally expensive and sensitive to the performance of auxiliary components such as autoencoders. Its effectiveness also depends on the nature and structure of the dataset being used.
 
\textbf{Learning to Explain (L2X)} is an information-theoretic method that explains model predictions by selecting input features that maximize mutual information with the output \cite{chen2018learning}. It trains an explainer network to identify the most informative feature subset, making it broadly applicable across architectures. However, its performance depends on the quality and representativeness of the training data, which can affect the relevance of selected features.

\textbf{Adversarial Black Box Explainer Generating Latent Exemplars (ABELE)} interprets model decisions by generating representative latent-space exemplars via adversarial perturbations and approximating the local decision boundary with a decision tree \cite{guidotti2020black}. It provides interpretable, example-based explanations aligned with the model's internal representations. However, it incurs substantial computational overhead due to the combined costs of latent space generation and surrogate model training, particularly when applied to high-dimensional image data.

\textbf{Training Calibration-Based Explainers (TraCE)} is tailored for medical image analysis and generates counterfactual explanations based on model calibration. It integrates uncertainty calibration into the model training process using the Learn-by-Calibrating (LbC) framework \cite{thiagarajan2020designing}, which adjusts output probabilities to ensure that predictions are both accurate and accompanied by well-calibrated uncertainty estimates, thereby improving the reliability of counterfactual explanations. However, it has high computational complexity, is dependent on the quality of the autoencoder, and its effectiveness may vary depending on the dataset and task.

\textbf{Explanation via Influence Functions} is adapted from robust statistics and assesses how individual training samples affect model predictions by approximating the impact of upweighting a sample on model parameters and, consequently, on the prediction outcome \cite{koh2017understanding}. Influence functions offer a principled approach to attribution without requiring retraining, making this method applicable to black-box settings. However, its effectiveness relies on strong assumptions like model differentiability and convexity, which are often not satisfied in modern deep neural networks. Moreover, it can be computationally infeasible for large-scale models due to the need for approximating inverse Hessians.

\subsubsection{Concept-Based Methods} 
\textbf{Concept Bottleneck Models (CBMs)} explain predictions by introducing an intermediate layer of human-interpretable concepts (e.g., ``narrow joint space"), separating the DL process into concept prediction followed by classification or regression \cite{koh2020concept}. This design enables users to trace, manipulate, and evaluate the conceptual basis of model decisions (e.g., ``Would the model still predict arthritis without joint space narrowing?"). While well-aligned with clinical reasoning, CBMs require labor-intensive concept-level annotations, and their interpretability and reliability depends on the quality and relevance of these annotations.

\textbf{Post-hoc CBMs (PCBMs)} addresses key limitations of CBMs by reducing reliance on dense concept annotations and preserving predictive performance \cite{yuksekgonul2022post}. They enable concept transfer from external datasets or natural language to reduce the annotation burden, and incorporate residual modeling to retain accuracy while providing concept-level interpretability. However, their effectiveness depends on the quality of the concept library, and the added architectural complexity may hinder transparency and deployment.

\textbf{Probabilistic CBMs (ProbCBMs)} extends traditional CBMs by modeling concepts as probability distributions rather than deterministic labels, enabling uncertainty quantification in concept representations \cite{yuksekgonul2022post}. This enhances robustness and reduces reliance on perfectly annotated concept labels. However, performance remains sensitive to concept supervision quality, and the added computational complexity may hinder scalability in large-scale applications.

\textbf{Testing with Concept Activation Vectors (TCAVs)} quantifies the influence of human-interpretable concepts on model predictions by analyzing their alignment in the network’s activation space \cite{kim2018interpretability}. Given user-defined concepts and counterexamples, it trains a linear classifier to derive a Concept Activation Vector (CAV), then measures the model’s sensitivity to perturbations along this direction. While TCAV provides global, semantically meaningful explanations, its reliability depends on the quality and specificity of the defined concepts.

\textbf{Automatic Concept-Based Explanation (ACEs)} automatically discovers and quantifies human-interpretable concepts by segmenting input images at multiple resolutions and embedding the segments into model’s activation space \cite{ghorbani2019towards}. Clustering is applied to identify distinct concepts, whose importance is then evaluated using TCAV. ACE eliminates the need for manual concept definition, offering semantically grounded explanations. However, its effectiveness depends on segmentation and clustering quality, and it incurs high computational cost, especially on large datasets.

\textbf{Visual Concept Mining (VCM)} provides semantically meaningful explanations for fine-grained classification tasks by identifying clinically relevant regions via segmentation and saliency-guided refinement \cite{fang2020concept}. These regions are clustered using self-supervised learning to form visual concepts, whose influence on model predictions is assessed through sensitivity analysis (e.g., TCAV). While VCM improves both interpretability and predictive performance, its effectiveness depends on segmentation and clustering quality and involves significant computational cost.

\textbf{ConceptSHAP} extends TCAV and ACE by introducing ``completeness", the degree to which a set of high-level concepts explains a model’s predictions \cite{yeh2020completeness}. It clusters intermediate activations to extract concept vectors, evaluates their predictive sufficiency via a completeness score, and quantifies individual concept contributions using Shapley values. ConceptSHAP provides global, semantically grounded explanations and supports automated concept discovery across modalities. However, it is computationally intensive and sensitive to both concept quality and model architecture.

\textbf{Causal Concept Effect (CaCE)} quantifies the causal influence of human-interpretable concepts on model predictions by estimating the effect of concept-level interventions, rather than relying on correlations \cite{goyal2019explaining}.  It provides more faithful, category-level explanations and is model-agnostic. However, it is computationally intensive due to the need for generating numerous contrastive samples to assess causal effects.

\subsubsection{Text-Based Methods}
\textbf{Visual Question Answering (VQA)} serves as an XAI technique by combining visual and textual modalities to generate interpretable, context-aware explanations \cite{antol2015vqa}. Given an input image and a related natural language question, the VQA model integrates visual features with semantic cues to produce an answer, revealing which regions or attributes inform the prediction. For example, the model may localize tumor boundaries to answer tumor size-related queries in CT scans. This modality-aligned reasoning enhances interpretability, particularly valuable in complex or ambiguous medical image analysis. However, its effectiveness depends on the quality and scope of training data, the clarity of questions, and the significant computational demands. 

\textbf{Image Captioning} serves as an XAI technique by generating natural language descriptions that summarize image content, including template-based, retrieval-based, and neural network-based methods \cite{Ayesha2021}. The former rely on predefined sentence structures or similar annotated examples, while neural models, which combine CNN visual encoders with RNN or Transformer language decoders, offer greater flexibility and explanatory depth. By aligning visual and linguistic representations, this method provides semantically rich, context-aware explanations beyond class labels, capturing object attributes, interactions, and scene-level context. However, its effectiveness depends on the quality and diversity of training data and may produce vague or inaccurate descriptions in complex medical images.

\textbf{Image Captioning with Visual Attention} extends traditional image captioning by incorporating attention mechanisms, enabling the model to focus on salient image regions when generating each word in the caption \cite{xu2015show}. It enhances interpretability by spatially aligning visual features with textual tokens, providing grounded explanations of model reasoning. While it enhances the descriptive precision compared to global-context captioning methods, it also introduces increased computational overhead and requires large-scale annotated datasets for effective training.

\textbf{Visual Dialog} is an interactive XAI method that explains model predictions through multi-turn conversations integrating vision and language \cite{das2017visual}. Unlike traditional VQA that handle isolated queries, it maintains contextual coherence across sequential questions by leveraging both image features and dialogue history, enabling more nuanced and iterative interpretation. This is particularly valuable in complex tasks such as medical image analysis, where iterative inquiry can uncover detailed rationale behind predictions. However, it requires substantial computational resources, high-quality annotated data, and robust dialogue understanding to ensure coherence and relevance.

\subsection{Latent-Based Methods} 
Latent-based methods interpret DL models by analyzing their internal representations in a reduced-dimensional latent space. By revealing underlying structures and feature dependencies, these methods provide insights into how learned representations influence model predictions.

\textbf{T-distributed Stochastic Neighbor Embedding (t-SNE)} projects high-dimensional feature representations into low-dimensional space by preserving local similarities, enabling visual analysis of clusters and class separability within learned representations. It is widely used for understanding the internal structure of DL networks in biomedical imaging tasks such as disease classification and lesion detection \cite{Nafisah2022,hussain2022shape,de2023explainable}. Despite its utility, t-SNE has limitations, including high computational cost, sensitivity to hyperparameters, and limited ability to preserve global data structure, which may affect the stability and reproducibility of its visualizations.

\textbf{Uniform Manifold Approximation and Projection (UMAP)} is a manifold learning-based dimensionality reduction technique that preserves both local and global structures more effectively than t-SNE \cite{mcinnes2018umap}. It constructs a weighted graph to approximate the low-dimensional manifold, enabling scalable and efficient embeddings. UMAP has been widely applied in biomedical imaging tasks such as breast lesion and cardiac amyloidosis classification \cite{hussain2022shape, de2023explainable}. However, it is sensitive to hyperparameters (e.g., number of neighbors, minimum distance), and may yield inconsistent results due to initialization and randomness. Its explanatory power may also be limited in highly nonlinear scenarios.

\section{Applications of XAI Across Biomedical Imaging Modalities} \label{sec3}
To contextualize XAI in biomedical image analysis, this section reviews its applications across major imaging modalities. By organizing studies by modality, we highlight how XAI methods address modality-specific characteristics, diagnostic objectives, and interpretability needs. For each modality, representative works are summarized and categorized by XAI method class, clarifying prevalent approaches and their alignment with clinical and technical requirements.

\begin{table}[htbp]
\centering
\caption{Representative XAI applications in radiographic image analysis: modalities, methods, and tasks.}
\label{tab:xai_radiographic}
\vspace{-0.2cm}
\renewcommand{\arraystretch}{1}
{
\resizebox{0.9\textwidth}{!}{
\begin{tabular}{llll}
\toprule[1pt]
\textbf{\makecell[l]{Imaging Modality}}  & \textbf{XAI Method} & \textbf{Reference} & \textbf{Task} \\
\midrule[1pt]
\multirow{31}{*}{CXR}  & CAM & \cite{Dunnmon2019} & Chest radiograph classification\\
  & Grad-CAM/++, Ablation-CAM & \cite{Hroub2024} & Lung disease prediction\\
  & Grad-CAM & \cite{Islam12023} & Lung disease  classification \\
  & Grad-CAM & \cite{Sheu2023} & Pneumonia infection classification \\
 &Grad-CAM, LRP & \cite{ikechukwu2023copdnet} & COPD diagnosis\\
  & Grad-CAM, t-SNE & \cite{Nafisah2022} & Tuberculosis detection and classification  \\
 &SHAP, LIME, Grad-CAM & \cite{Bhandari2022} & Pneumonia and Tuberculosis classification \\
  &Grad-CAM & \cite{rani2022multi} & lung cancer classification  \\
 & Grad-CAM,LIME & \cite{Ali2022} & COVID-19 detection and classification \\
&Grad-CAM & \cite{Sharma2022} & Severity assessment and diagnosis of COVID-19 \\
 & Grad-CAM & \cite{Singh2021} & COVID-19 and Pneumonia classification \\
  & Grad-CAM & \cite{chetoui2021deep} & COVID-19 detection and classification \\
 & Grad-CAM++, LRP & \cite{karim2020deepcovidexplainer} & COVID-19 detection and classification \\
 & Grad-CAM & \cite{Brunese2020} & COVID-19 detection and classification \\
 & Grad-CAM & \cite{Lee2020} & COVID-19 screening and classification \\
  & Grad-CAM & \cite{Mahmud2020} & COVID-19 and Pneumonia detection, and classification \\
 & LRP & \cite{pitroda2021explainable} & lung disease classification\\
  & LIME & \cite{Koyyada2023} &  lung disease  detection and classification \\
  & LIME, SHAP & \cite{ong2021comparative} & COVID-19 and Pneumonia classification  \\
 & LIME & \cite{Ahsan2021} & COVID-19 detection and classification \\
  & Attention, Grad-CAM & \cite{Ukwuoma2023} & Pneumonia classification\\
 & Image Captioning & \cite{gajbhiye2020automatic} & Automatic report generation \\
  & Image Captioning & \cite{singh2022efficient} & Automatic report generation \\
   & Image Captioning & \cite{kayser2022explaining} & Explaining CXR pathologies \\
  & Image Captioning & \cite{rio2024parameter} & Explanations generation for CXR classification \\
  & Image Captioning & \cite{kaur2022radiobert} & Automatic report generation \\
  & Score-CAM,t-SNE & \cite{Rahman2020} & Tuberculosis segmentation and classification \\
  & VQA &\cite{cong2022caption} & Medical question answering \\
\hline
\multirow{10}{*}{Mammography}  & Eigen-CAM & \cite{prinzi2024yolo} & Breast cancer detection \\
  & Grad-CAM & \cite{farrag2023explainable} & Mammogram tumor segmentation \\
  & Attention Grad-CAM & \cite{raghavan2023attention} &Breast cancer detection and classification \\
  & Deep SHAP, Grad-CAM & \cite{gerbasi2023deepmica} &Breast microcalcification malignancy detection and classification \\
& Grad-CAM & \cite{kang2021convolutional} & Breast microcalcification classification\\
  & Grad-CAM & \cite{suh2020automated} & Beast cancer detection\\
  & Grad-CAM,LIME & \cite{ahmed2024coatnet} & Mammogram mass detection and classification \\
  &Grad-CAM &\cite{pertuz2023saliency} &Beast cancer detection and classification\\
  & Image captioning & \cite{pang2024radioport} & Report generation for mammographic calcification classification\\
   & Image captioning & \cite{luong2024toward} & Report generation for breast cancer diagnosis\\ \hline
\multirow{2}{*}{\makecell[l]{Digital \\Tomosynthesis}} & Grad-CAM & \cite{ricciardi2021deep} & Breast tomosynthesis lassification\\
 & Grad-CAM,LIME,t-SNE,UMAP & \cite{hussain2022shape} & Breast lesion classification\\
\hline
\multirow{2}{*}{Fluoroscopy}  & Grad-CAM & \cite{zhang2024multi} &Vertebral compression fractures segmentation \\
           & Grad-CAM & \cite{terunuma2023explainability} & Tumor egmentation \\
\bottomrule[1pt]
\end{tabular}
}}
\end{table}

\subsection{XAI in Radiographic Image Analysis}
Radiographic modalities such as chest X-rays (CXR), mammography, and fluoroscopy are widely used in clinical workflows due to their efficiency, accessibility, and diagnostic utility. Among them, CXR has become a benchmark for DL-based disease classification, particularly for detecting pulmonary conditions like tuberculosis, pneumonia, COVID-19, and chronic obstructive pulmonary disease. Given their direct impact on clinical decisions, the need for interpretable models in this domain is critical. Visual explanation methods, especially Grad-CAM and its variants, have dominated this space, offering intuitive heatmap overlays to highlight salient regions. For example, \cite{karim2020deepcovidexplainer} integrated Grad-CAM, Grad-CAM++, and LRP to explain predictions for COVID-19 and pneumonia. Similarly, \cite{pitroda2021explainable} compared LIME, guided backpropagation, and LRP for lung disease interpretation. These methods also extend beyond binary classification. For example, Grad-CAM has been applied to tuberculosis detection \cite{Nafisah2022}, and LIME-enhanced CNNs were used for interpretable COVID-19 diagnosis \cite{Koyyada2023}. SHAP-based methods have also gained prominence. For instance, \cite{Zou2023} proposed an ensemble approach combining SHAP and Grad-CAM, outperforming traditional saliency-based methods. Beyond thoracic imaging, visual explanation techniques have been extended to breast lesion classification \cite{hussain2022shape} and microcalcification analysis \cite{gerbasi2023deepmica}, demonstrating their versatility across radiographic tasks.

While visual explanations dominate, text-based methods have emerged as promising complements. \citet{gajbhiye2020automatic} used an image captioning model to generate descriptive summaries from CXR scans, bridging the gap between visual evidence and clinical reasoning. Despite this progress, current XAI approaches are largely post-hoc and often lack clinical validation, robustness across populations, or resistance to spurious correlations. Grad-CAM and similar methods are also sensitive to model architecture and may highlight irrelevant regions. Moving forward, integrating visual, textual, and example-based explanations with uncertainty quantification and user feedback will be key to building trustworthy radiographic AI systems. Table~\ref{tab:xai_radiographic} summarizes representative XAI applications across radiographic imaging.

\subsection{XAI in Computed Tomography (CT) Image Analysis}
CT imaging is essential for diagnosing thoracic and neurological conditions, and XAI techniques have been increasingly integrated to improve interpretability in DL-based CT analysis \cite{Pennisi2021}. Most studies rely on attribution-based methods, particularly CAM variants such as Grad-CAM, to generate voxel- or region-level heatmaps aligned with radiological findings. For example, Grad-CAM was applied in Lung-EFFNet to localize malignancy-relevant areas in lung cancer classification \cite{raza2023lung}, while Grad-CAM and LIME were jointly used for interpretable COVID-19 diagnosis \cite{ye2021explainable}. A comparative study by \cite{lima2022evaluation} further evaluated the consistency of CAM-based explanations. Hybrid architectures combining CNNs and GRUs have also incorporated LIME, SHAP, and Grad-CAM to enhance interpretability in lung disease analysis \cite{Islam2023}. While effective for highlighting spatially discriminative regions, these methods often provide coarse localization, are architecture-sensitive, and may not reflect causal features.

To complement visual attribution, text-based methods have been explored to enhance semantic interpretability. Image captioning models have been used to describe key features in CT scans, offering clinician-friendly, natural language explanations. For example, \cite{magalhaes2024image} utilized captioning for coronary artery disease interpretation, and \cite{kim2023convolutional} adopted captioning for intracerebral hemorrhage detection. These methods support human-AI collaboration but depend on high-quality annotations and robust language generation, which can introduce ambiguity or overlook critical features.

Despite these advances, XAI in CT imaging remains largely focused on post-hoc explanations. There is growing need for concept-based and counterfactual approaches that go beyond spatial attribution to offer more actionable, causally grounded insights. Additionally, explanation fidelity is often under-evaluated, with limited alignment validation between model rationales and clinical reasoning. Future research should prioritize clinically meaningful evaluation metrics, causal interpretability techniques, and interactive explanation interfaces to build trustworthy, decision-supportive AI systems. Table~\ref{tab:xai_ct} summarizes representative studies categorized by explanation type and clinical application.

\begin{table}[htbp]
\caption{Representative XAI applications in CT image analysis: modalities, methods, and tasks.}
\label{tab:xai_ct}
\centering
\vspace{-0.2cm}
\resizebox{0.85\textwidth}{!}{
\begin{tabular}{llll}
\toprule[1pt]
\textbf{\makecell[l]{Imaging Modality}}  & \textbf{XAI Method} & \textbf{Reference} & \textbf{Task} \\
\midrule[1pt]
\multirow{19}{*}{CT} & CAM & \cite{lima2022evaluation} & COVID-19 classification \\
 & Grad-CAM & \cite{Mercaldo2023} & COVID-19 detection\\
& Grad-CAM & \cite{Pennisi2021} & COVID-19 assessment and lesion classification \\
  & LIME & \cite{Beddiar2023} & COVID-19  detection and classification \\
  & LIME, Grad-CAM & \cite{ye2021explainable} & COVID-19 CT classification \\
& Grad-CAM, LIME, SHAP & \cite{Islam2023} & lung abnormalities detection and classification \\
  &Grad-CAM & \cite{raza2023lung} & Lung cancer classification \\
  & LIME & \cite{Ahsan2021} & COVID-19 detection and classification \\
  & HiRe-CAM , Grad-CAM & \cite{draelos2022explainable} & Chest abnormality classification \\
    & Grad-CAM++ & \cite{li2022explainable} &  COVID-19 detection and lesion segmentation  \\
      & Grad-CAM, Guided Grad-CAM & \cite{niranjan2023explainable} &  COVID-19 classification and segmentation \\
      &Grad-CAM++ & \cite{atim2024explainable} &  Ground glass opacities segmentation \\
       & Grad-CAM, LIME, IG & \cite{elbouknify2023ct} & COVID-19 detection and segmentation \\
        & Grad-CAM,Grad-CAM++ & \cite{darapaneni2022explainable} & COVID-19 and lesion segmentation and classification \\
       & Grad-CAM & \cite{islam2022vision} & Kidney cyst, stone and tumor detection \\
           &LIME, Grad-CAM            & \cite{yildirim2023image} & Hydatid cysts classification \\
        & Prototype & \cite{soares2024explainable} & COVID-19  classification and segmentation \\  
 & Image captioning &\cite{magalhaes2024image} & Coronary artery disease diagnosis \\
  & Image captioning &\cite{kim2023convolutional} & Intracranial hemorrhage diagnosis \\
& Image captioning &\cite{tang2024work} & Report generation \\
& Image captioning &\cite{kim2024ftt} & Brain CT report generation \\
  & VQA &\cite{cong2022caption} & Medical question answering \\
\hline
  Cone Beam CT &Grad-CAM++ & \cite{barzas2023explainable} & Mandibular canal segmentation \\
\bottomrule[1pt]
\end{tabular} }
\end{table}

\subsection{XAI in Magnetic Resonance Imaging (MRI) Image Analysis}
MRI and its variants, functional MRI (fMRI), magnetic resonance angiography (MRA), and spectroscopy (MRS), are central to the diagnosis of neurological and neurovascular disorders. In this domain, XAI has been increasingly adopted to enhance interpretability, clinical trust, and understanding of model behavior.
Alzheimer’s disease (AD) classification is a major application area, where saliency and attribution techniques have been widely used. LIME and LRP have been applied in multimodal models combining MRI and genetic data, while SHAP has enabled detailed feature attribution in multimodal AD frameworks \cite{mahim2024unlocking}.

Grad-CAM has been employed to localize disease-relevant brain regions in structural MRI \cite{mahmud2024explainable}, with an emphasis on alignment with neuroanatomical knowledge. Beyond classification, XAI has also supported segmentation and detection tasks. \citet{zeineldin2022explainability} evaluated multiple techniques (e.g., Integrated Gradients, SmoothGrad) for surgical decision support. In tumor segmentation, models such as NeuroNet19 \cite{haque2024neuronet19} and LIME-integrated ensembles \cite{10100703} provided both spatial and feature-level explanations. High-resolution attention mechanisms have also been benchmarked against Grad-CAM for improved localization precision \cite{yu2022novel}. XAI has further been extended to vascular and functional imaging. For example, Grad-CAM was used in MRA-based moyamoya disease detection \cite{yin2022magnetic}, and ViT-GRU models combined attention and SHAP for interpretability in fMRI-based diagnosis \cite{mahim2024unlocking}.

While these advances demonstrate the growing maturity of XAI in magnetic imaging, several challenges remain. Many methods lack robustness across imaging protocols, scanners, and patient populations. Most existing work is post-hoc, with limited integration of anatomical priors or domain constraints. Functional imaging methods, such as those for fMRI, still struggle with temporal interpretability and causal alignment. Future research should explore concept-based and counterfactual explanations, uncertainty-aware interpretability for time-series data, and standardized benchmarks for validation. Table~\ref{tab:xai_mri} summarizes representative studies categorized by modality, task, and XAI technique.

\begin{table}[htbp]
\caption{Representative XAI applications in MRI image analysis: modalities, methods, and tasks.}
\label{tab:xai_mri}
\centering
\vspace{-0.2cm}
\resizebox{0.75\textwidth}{!}
{
\begin{tabular}{llll}
\toprule[1pt]
\textbf{\makecell[l]{Imaging Modality}}  & \textbf{XAI Method} & \textbf{Reference} & \textbf{Task} \\
\midrule[1pt]
\multirow{17}{*}{MRI} & Grad-CAM & \cite{mahmud2024explainable} & Alzheimer’s disease(AD) classification \\
 & Grad-CAM & \cite{natekar2020demystifying} & Brain tumor segmentation and classification\\
 & Grad-CAM & \cite{dasanayaka2022interpretable} & Brain tumor segmentation and classification\\
 & Grad-CAM & \cite{farhan2025xai} & 3D brain tumor segmentation and classification\\
 & Grad-CAM & \cite{t2024enhancing} & Brain tumor detection and classification\\
 & Grad-CAM,Grad-CAM++ & \cite{nhlapho2024bridging} & Brain tumor detection and classification\\
 & Grad-CAM,Grad-CAM++ & \cite{hussain2023explainable} & Tumor classification and localization\\
  & LRP & \cite{shin2023deep} & Tumor segmentation and classification \\
  & LRP & \cite{mandloi2024explainable} & Brain tumor detection and classification\\
  & LIME & \cite{haque2024neuronet19} & Brain tumors classification \\
  & SHAP & \cite{ahmed2023effective} & Brain tumor detection and classification \\
  & LIME& \cite{lakshmi2025explainable} & Brain tumor segmentation and classification\\
  & LIME & \cite{10100703} & Brain tumor detection and classification\\
  & LIME & \cite{ullah2024enhancing} & Brain tumor detection and classification\\
 & Image-Captioning & \cite{mayzura2025automatic} & Automatic brain image interpretation \\
 & Attention,LIME,SHAP & \cite{mahim2024unlocking} & AD detection and classification \\
   & ProtoPNet & \cite{wei2024mprotonet} & Brain tumor classification \\ \hline
MRA  & Grad-CAM & \cite{yin2022magnetic} & Moyamoya disease classification \\ \hline
\multirow{2}{*}{fMRI}  & LRP & \cite{ellis2023towards} & Schizophrenia diagnosis and classification\\
  & Grad-CAM & \cite{song2024explainability} & AD classification \\ \hline
fMRI,MRI & Grad-CAM & \cite{varaprasad2025exploring} &Schizophrenia classification \\
\bottomrule[1pt]
\end{tabular} }
\end{table}

\subsection{XAI in Ultrasound Image Analysis}
Ultrasound and elastography are widely used for real-time, point-of-care diagnostics obstetrics, cardiology, hepatology, and oncology. However, interpretation is often hampered by variability in acquisition quality, probe angle, and operator expertise. XAI methods in this domain aim to address these modality-specific limitations.
Visual attribution techniques, particularly heatmap-based methods like Grad-CAM, have been extensively applied to highlight salient regions in fetal biometry, cardiac imaging, and thyroid nodule classification \cite{Zhang2023}. These methods provide intuitive, real-time visual cues that support bedside decision-making. Perturbation-based techniques such as LIME and SHAP offer finer-grained explanations by quantifying feature contributions, as demonstrated in liver fibrosis staging and cataract grading. Beyond visual explanations, generative and textual approaches are gaining traction. \citet{alsharid2019captioning} proposed an image captioning framework to generate radiology-style reports from ultrasound images. Similarly, \citet{rezazadeh2022explainable} developed a multimodal model integrating SHAP-based attribution with natural language explanations for breast cancer detection. ThyExp \cite{Morris2023}, an interactive web-based system, combines interpretable AI with clinician-facing visualizations for thyroid imaging, showcasing the potential for practical XAI integration.

Despite these advances, ultrasound-specific challenges remain. Operator dependence introduces variability that hinders model generalization and complicates explanation benchmarking. Most current methods are post-hoc and qualitative, with limited alignment to expert annotations or clinical outcomes. Additionally, concept-based and counterfactual explanations tailored to ultrasound pathologies are largely underexplored. Future work should focus on quantitative evaluation frameworks, domain-adaptive explanation strategies, and user-centered design to promote broader adoption. Table~\ref{tab:xai_ultrasound} summarizes XAI applications in ultrasound, categorized by method type and analysis task.

\begin{table}[htbp]
\caption{Representative XAI applications in ultrasound image analysis: modalities, methods, and tasks.}
\label{tab:xai_ultrasound}
\centering
\vspace{-0.2cm}
\resizebox{0.8\textwidth}{!}
{
\begin{tabular}{llll}
\toprule[1pt]
\textbf{\makecell[l]{Imaging Modality}} & \textbf{XAI Method} & \textbf{Reference} & \textbf{Task} \\
\midrule[1pt]
\multirow{16}{*}{Ultrasound~~~}   & Grad-CAM            & \cite{Zhang2023} & Fetal congenital heart disease classification \\
    & Grad-CAM            & \cite{duque2024ultrasound} & ultrasound image segmentation \\
      & Grad-CAM            & \cite{singh2024atherosclerotic} & Atherosclerotic plaque classification \\
      & CAM            & \cite{luo2022segmentation} & Breast tumor classification \\
      & CAM            & \cite{yang2022ctg} & Breast ultrasound segmentation and classification \\
       & CAM            & \cite{yu2022adaptive} & Thyroid ultrasound segmentation \\
       & Grad-CAM            & \cite{islam2024enhancing} & Breast cancer segmentation and classification \\
        & Grad-CAM            & \cite{ho2022classification} & Rotator cuff tears classification \\
        & Grad-CAM            & \cite{katakis2023muscle} & Muscle cross-sectional area classification and segmentation \\
       & Grad-CAM            & \cite{butt2024beyond} & Breast cancer detection and classification \\
       & Grad-CAM            & \cite{kang2022thyroid} & Thyroid nodule segmentation and classification\\
      & SHAP & \cite{hasan2023fp} & ultrasound  lung classification \\
      & LIME & \cite{harikumar2024explainable} & Fetal ultrasound classification \\
       & Attention & \cite{manh2022multi} & Thyroid cancer classification \\
           & Image Captioning & \cite{alsharid2019captioning} & Report generation  \\
    & Image captioning & \cite{luong2024toward} & Report generation for breast cancer diagnosis\\ 
\bottomrule[1pt]
\end{tabular}
}
\end{table}

\subsection{XAI in Positron Emission Tomography (PET) Image Analysis}
Although still emerging, the integration of XAI into nuclear imaging, particularly PET and SPECT, is showing strong potential to enhance interpretability in functional diagnostics \cite{yang2025ai}.
These modalities capture metabolic and molecular activity critical for diagnosing neurological, oncological, and cardiovascular diseases, where voxel-level interpretability is especially valuable.
Saliency-based techniques have been widely used to localize functionally relevant biomarkers. For example, \citet{nazari2022explainable} applied Layer-wise Relevance Propagation (LRP) to 3D CNNs for visualizing striatal uptake patterns in DaTscan SPECT, aiding early Parkinson’s diagnosis. Similarly, \citet{jiang2025using} combined SHAP and Grad-CAM to identify disease-relevant regions in Alzheimer's prediction using PET scans. Latent space visualizations have also been employed to interpret model behavior. For example, \citet{de2023explainable} used t-SNE, UMAP, and related methods to reveal phenotype clustering in cardiac amyloidosis. In multimodal contexts, \citet{jiang2025explainable} proposed an interpretable PET-clinical fusion model for follicular lymphoma prognosis, using SHAP and Grad-CAM to assess contributions from both imaging and clinical features.

Despite promising progress, XAI in nuclear imaging faces key challenges. Annotated PET/SPECT datasets remain limited, image quality is affected by high noise levels, and explanation methods often lack clinical validation. Current methods largely focus on visualization, with limited evaluation of whether explanations align with radiological reasoning. Future research should emphasize method robustness, cross-modal consistency, and integration with expert feedback to ensure explanations are not only interpretable but also clinically actionable. Table~\ref{tab:xai_nuclear} summarizes representative XAI applications in PET and SPECT imaging, categorized by method type and diagnostic task.

\begin{table}[htbp]
\caption{Representative XAI applications in PET image analysis: modalities, methods, and tasks.}
\label{tab:xai_nuclear}
\centering
\vspace{-0.2cm}
\resizebox{0.7\textwidth}{!}
{
\begin{tabular}{llll}
\toprule[1pt]
\textbf{Imaging Modality}  & \textbf{XAI Method} & \textbf{Reference} & \textbf{Task} \\
\midrule[1pt]
\multirow{6}{*}{SPECT} & LRP & \cite{nazari2022explainable} & Dopamine transporter SPECT classification \\
      & Grad-CAM & \cite{papandrianos2022explainable} & Coronary artery disease lassification \\
       & Grad-CAM & \cite{chen2021convolutional} & Myocardial perfusion images classification \\
      & Grad-CAM & \cite{otaki2022clinical} & Coronary artery disease detection \\
      & Grad-CAM & \cite{kusumoto2024deep} & Coronary artery disease diagnosis \\
      & Attention & \cite{thakur2022soft} & Parkinson’s disease classification \\ \hline
\multirow{3}{*}{PET} & UMAP, t-SNE & \cite{de2023explainable} & Cardiac Amyloidosis classification \\
       & SHAP, Grad-CAM & \cite{jiang2025using}& Early AD spectrum prediction\\
        & SHAP & \cite{duan2025non}& Lymph node metastasis prediction\\
\bottomrule[1pt]
\end{tabular}}
\end{table}

\subsection{XAI in Optical Image Analysis}
Optical imaging modalities, including dermoscopy, fundus photography, and optical coherence tomography (OCT) are widely used in dermatology and ophthalmology for non-invasive, high-resolution visualization of skin and ocular structures. As DL models are increasingly adopted in these fields, explainability has become essential for building clinician trust and support diagnostic decision-making.

\textbf{Dermatology.} In skin lesion analysis and cancer detection, XAI efforts have primarily focused on enhancing visual interpretability of CNN-based classifiers. CAM-based techniques, particularly Grad-CAM and its variants, are the most widely adopted. For example, DermX \cite{Jalaboi2023} and other Grad-CAM-integrated models \cite{mayanja2023explainable, Mridha2023, ZiaUrRehman2022, nunnari2021overlap} have demonstrated improved lesion localization and model transparency. Grad-CAM++ and Eigen-CAM were benchmarked against dermatologist annotations \cite{GiavinaBianchi2023}. Perturbation-based methods such as LIME and SHAP have also been applied \cite{stieler2021towards, Nigar2022, Khater2023}, though they often lack spatial precision. Hybrid strategies combining multiple explanation techniques have been proposed to enhance robustness \cite{athina2022multi, Shorfuzzaman2021}. In parallel, backpropagation- and attention-based methods, such as LRP \cite{Ganguly2022} and attention-enhanced CNNs \cite{Barata2021}, are gaining popularity. Concept-level explanations are also emerging, including ACE \cite{sauter2022validating} and multimodal frameworks integrating visual and textual outputs \cite{Lucieri2022}.

\textbf{Ophthalmology.} In retinal and OCT image analysis, XAI has supported early detection of conditions such as diabetic retinopathy and glaucoma. Visual attribution remains dominant, with Grad-CAM frequently used to localize pathological features \cite{jiang2020multi, vasquez2022interactive}, often combined with SHAP, LIME, or guided Grad-CAM for enhanced interpretability \cite{vieira2023applied,reza2021interpretable, volkov2023possibilities}. Attention-based models, especially Transformer-based models like Focused Attention \cite{playout2022focused}, have shown promise by aligning attention maps with clinical regions of interest, improving explainability over traditional CNNs.

Across dermatology and ophthalmology, XAI is evolving toward multimodal, user-centered, and clinically meaningful interpretability. Key challenges remain, including variation in image acquisition, class imbalance, and the lack of standardized metrics for evaluating explanation quality. Future work should integrate concept-level reasoning, human-in-the-loop validation, and cross-modal interpretability to enhance the reliability and clinical utility of AI systems. Table~\ref{tab:xai_optical} summarizes representative XAI applications in optical imaging, organized by method type and clinical task.

\begin{table}[htbp]
\caption{Representative XAI applications in optical image analysis: modalities, methods, and tasks.}
\label{tab:xai_optical}
\centering
\vspace{-0.2cm}
\resizebox{0.85\textwidth}{!}
{
\begin{tabular}{llll}
\toprule[1pt]
\textbf{\makecell[l]{Imaging Modality}} & \textbf{XAI Method} & \textbf{Reference} & \textbf{Task} \\
\midrule[1pt]
\multirow{13}{*}{Dermatology} & Grad-CAM & \cite{Jalaboi2023} & Skin disease detection and classification \\
 & Grad-CAM & \cite{Ahmad2023} & Skin lesion recognition \\
  & Grad-CAM,Grad-CAM++ & \cite{Mridha2023} & Skin cancer classification\\
 & Grad-CAM & \cite{nunnari2021overlap} & Skin cancer classification\\
 & Grad-CAM & \cite{ZiaUrRehman2022} & Skin lesion classification\\
 & Grad-CAM, Smooth-Grad & \cite{mayanja2023explainable} & Skin disease classification \\
 & Grad-CAM & \cite{Barata2021} & Skin cancer classification\\
  & CAM & \cite{Zhang2019} & Skin lesion classification \\
 & LIME & \cite{Nigar2022} & Skin lesion classification \\
  & LIME & \cite{stieler2021towards} & Skin cancer classification \\
  & SHAP & \cite{Khater2023} & Skin cancer classification\\
 & TCAV/CBM & \cite{Lucieri2022} & Skin lesions diagnosis\\
  & TCAV/CBM & \cite{patricio2023coherent} & Skin lesions diagnosis\\ \hline
\multirow{6}{*}{OCT}  & Grad-CAM, occlusion,LIME & \cite{Han2024} & Synthesize OCT images for eye diagnosis\\
 & Grad-CAM & \cite{Bui2023} & Retinal diseases detection and classification\\
  & Grad-CAM & \cite{vasquez2022interactive} & Retinal disease classification\\
 & LIME,SHAP & \cite{Bhandari2023} &Retinal disease classification \\
  & LIME, Grad-CAM & \cite{apon2021demystifying} & Retinal disease classification\\
  & LIME & \cite{reza2021interpretable} & Retinal disease classification\\ \hline
\multirow{6}{*}{Fundus} & Grad-CAM & \cite{Alghamdi2022} & Diabetic retinopathy detection and classification\\
    & Focused attention & \cite{playout2022focused} & Retinal image classification\\
  & Grad-CAM & \cite{Guo2021} & Eye diseases detection and classification\\
& Grad-CAM & \cite{jiang2020multi} &Diabetic retinopathy classification \\
& LIME & \cite{volkov2023possibilities} &Glaucoma detection detection and classification \\
  & Grad-CAM, Guided IG, XRAI & \cite{volkov2023gradient} & Eye disease classification\\ \hline
\multirow{6}{*}{Endoscopy}   & Grad-CAM & \cite{NoumanNoor2023} & Gastrointestinal tract disorders classification\\
 & Grad-CAM & \cite{mukhtorov2023endoscopic} & Endoscopic image classification\\
& Grad-CAM & \cite{garcia2022towards} & Endoscopic image classification\\
 & Grad-CAM & \cite{zheng2024enhancing} & Gastrointestinal submucosal tumor detection and lcassification\\
  &SHAP & \cite{auzine2023classification} & Gastrointestinal cancer classification\\
   & SHAP, Grad-CAM & \cite{ahamed2024detection}& Gastrointestinal tract diseases detection\\
\bottomrule[1pt]
\end{tabular} }
\end{table}

\subsection{XAI in Microscopy Image Analysis}
Microscopy imaging, including histopathology, cytology, confocal microscopy, and hematology, plays a crucial role in cellular-level disease diagnosis. The extreme resolution of whole-slide images (WSIs) poses major challenges for interpreting DL models. To address this, visual attribution methods such as Grad-CAM and HR-CAM have been extended to the tile level to highlight morphologically relevant features like mitotic figures, tumor margins, and lymphocytic infiltrates, especially in breast and head-and-neck cancer analysis \cite{Drrich2023}.
Perturbation-based methods such as LIME and SHAP have been adapted for cytology and gastrointestinal pathology to identify diagnostically meaningful regions or detect misleading model focus, thus aiding both transparency and quality assurance \cite{kaplun2021cancer, auzine2023classification}. In hematology, Grad-CAM variants have revealed key cytological traits, such as nuclear shape and granularity, influencing predictions in white blood cell classification and parasite detection \cite{islam2024explainable, mayrose2024explainable}.

Beyond pixel-level saliency, concept-based and attention-driven XAI methods are increasingly adopted. Methods such as ACE and prototype learning have enabled models to align internal features with expert-defined pathological concepts like ``keratin pearls" or ``nuclear pleomorphism" \cite{sauter2022validating}. Meanwhile, self-attention mechanisms and transformer-based architectures offer global interpretability by modeling long-range dependencies and generating clinically aligned attention maps. Examples include ESAE-Net, which integrates attention modules with XAI overlays for breast cancer classification \cite{peta2024explainable}, and attention rollout techniques applied to histopathology \cite{mir2025enhancing}. Despite these advances, challenges persist in standardizing explanations, validating clinical utility, and ensuring scalability for high-throughput deployment. Continued progress will depend on developing concept-aware, multi-resolution, and human-in-the-loop interpretability frameworks. Table~\ref{tab:xai_micro} summarizes representative XAI applications across microscopy modalities.

\begin{table}[h]
\caption{Representative XAI applications in microscopy image analysis: modalities, methods, and analysis tasks.}
\label{tab:xai_micro}
\centering
\vspace{-0.2cm}
\resizebox{0.85\textwidth}{!}
{
\begin{tabular}{llll}
\toprule[1pt]
\textbf{\makecell[l]{Imaging Modality}}  & \textbf{XAI Method} & \textbf{Reference} & \textbf{Task} \\
\midrule[1pt]
\multirow{17}{*}{Histopathology}  & Grad-CAM & \cite{yengec2023improved} & Colorectal polyps classification \\
  & Grad-CAM & \cite{civit2022non} & non-small cell lung cancer classification\\
 & Grad-CAM & \cite{qing2024mpsa} & Histopathological image classification\\
  & Grad-CAM,Guided Grad-CAM  & \cite{li2023convolutional} & Cervical cancer types classification\\
  & Grad-CAM & \cite{liu2023collaborative} & Breast cancer classification\\
  & Grad-CAM & \cite{liu2022contennet} & Breast cancer classification\\
  & Grad-CAM & \cite{macedo2022evaluating} & Breast cancer segmentation and classification\\
   & Grad-CAM,Guided Grad-CAM  & \cite{yang2022micl} & Histopathology image classification\\
   & Grad-CAM & \cite{yong2023histopathological} & Cancer detection and classification \\
   & Grad-CAM & \cite{diao2023deep} &  Histopathological image classification\\\
 & LIME & \cite{kaplun2021cancer} & Breast cancer classification \\
  & LRP, LIME, Attention rollout & \cite{mir2025enhancing} & Histopathological image classification \\
  & Attention & \cite{zhang2017mdnet} & Pathology bladder cancer diagnosis\\
& Grad-CAM, HR-CAM & \cite{Drrich2023} & Head and neck cancer segmentation and classification\\
   & Raw-attention & \cite{he2022deconv} & Breast cancer classification\\
  & ACE & \cite{sauter2022validating} & Validate automatic explanations for classification\\
 & Prototype & \cite{uehara2019prototype} &pathological image classification \\ \hline
\multirow{6}{*}{Hematology}  & Grad-CAM, Grad-CAM /++, LIME,SHAP & \cite{islam2024explainable} & Blood cell classification\\
  & Grad-CAM & \cite{goswami2024explainable} &  Sickle cell detection and classification\\
 & Grad-CAM & \cite{tsutsui2023wbcatt} & WBC classification\\
& Grad-CAM & \cite{mayrose2024explainable} &Dengue detection and classification \\
  & LIME & \cite{deshpande2024explainable} & Leukemia classification\\
  & CBM & \cite{pang2024integrating} & WBC classification\\
\bottomrule[1pt]
\end{tabular} }
\end{table}

\subsection{XAI in Multi-modality Biomedical Image Analysis}
The integration of imaging modalities such as MRI, PET, CT, and CXR offers complementary diagnostic information but poses unique challenges for interpretability, particularly in disentangling modality-specific contributions and aligning fused features with clinical reasoning \cite{jin2022evaluating}. Visual attribution techniques, especially Grad-CAM, have been widely adopted to interpret fused-model outputs. In neurodegenerative disease analysis, Grad-CAM has been used to highlight distinct structural and functional brain regions from MRI and PET inputs, supporting modality-specific interpretation in AD diagnosis \cite{liu2024hammf, zhou2022interpretable}.
In COVID-19 classification, CT and CXR were jointly analyzed using Grad-CAM to reveal modality-specific cues such as pulmonary texture and volumetric lung features \cite{el2024differences}. These studies illustrate the role of XAI in not only interpreting individual modalities but also in revealing their complementary diagnostic value.

Beyond classification, multimodal segmentation presents additional interpretability challenges. For example, \cite{zhao2024evidence} generated pixel-level attribution maps for multimodal segmentation, providing fine-grained insights into how distinct modalities shape spatial predictions. Despite these advances, current XAI methods lack mechanisms to quantify each modality’s contribution and rarely capture interactions between fused features. Furthermore, standard benchmarks for evaluating multimodal explanations remain underdeveloped. Future work should prioritize fusion-aware explanation strategies, cross-modality attribution methods, and interactive frameworks that support clinical decision-making at both global and local levels. Table \ref{tab:xai_multimodal} summarizes representative XAI applications in multimodal biomedical imaging.

\begin{table}[h]
\caption{Representative XAI applications in multi-modality biomedical image analysis: modalities, methods, and tasks.}
\label{tab:xai_multimodal}
\centering
\vspace{-0.2cm}
\resizebox{0.6\textwidth}{!}
{
\begin{tabular}{llll}
\toprule[1pt]
\textbf{Imaging Modality} & \textbf{XAI Method} & \textbf{Reference} & \textbf{Task} \\
\midrule[1pt]
\multirow{3}{*}{PET, MRI}  & Grad-CAM & \cite{liu2024hammf} & AD diagnosis \\
        & Grad-CAM & \cite{zhou2022interpretable} & AD diagnosis\\
        & Grad-CAM & \cite{castellano2024automated} & AD detection and classification\\
\hline
CT, CXR & Grad-CAM & \cite{el2024differences} & COVID-19 diagnosis\\
\hline
  PET, CT & LIME & \cite{yang2025multi}& Tumor segmentation\\
\hline
fMRI, sMRI & Score-CAM & \cite{zhou2024interpretable} & Schizophrenia diagnosis\\
\bottomrule[1pt]
\end{tabular} }
\end{table}

\subsection{Interpretable Vision–Language Models (VLMs) for Biomedical Image Analysis}
Recent advances in VLMs, particularly foundation models like CLIP \cite{radford2021learning}, have enabled cross-modal tasks such as zero-shot classification and semantic retrieval in biomedical imaging. Domain-specific variants, MedCLIP \cite{wang2022medclip} and BioMedCLIP \cite{zhang2023biomedclip}, adapt these models to medical semantics, showing promising results across radiology, pathology, ophthalmology, dermatology, and endoscopy.
However, most VLMs remain opaque, and conventional XAI methods (e.g., Grad-CAM, attention maps) fail to capture fine-grained multimodal reasoning, limiting clinical trust.

\begin{figure}[htbp]
  \centering
  \includegraphics[width=0.85\textwidth]{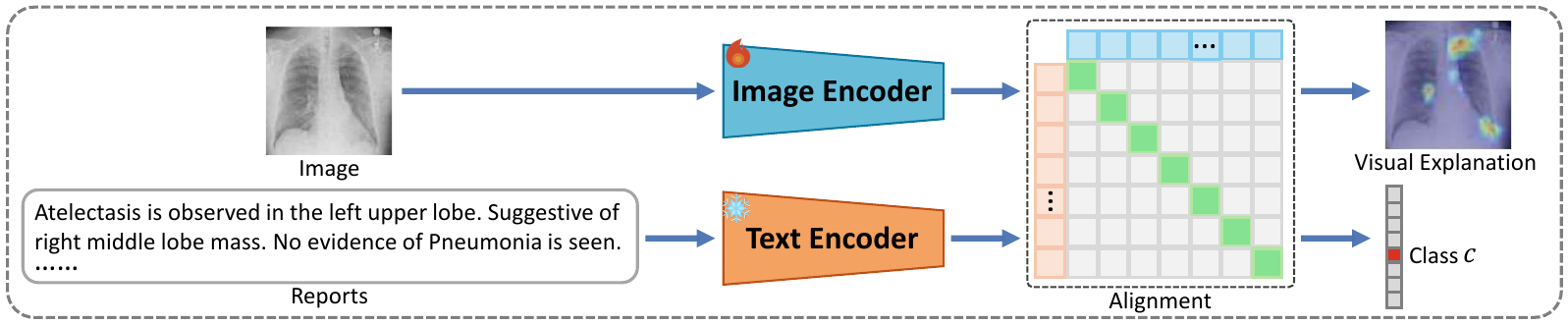}
  \vspace{-0.2cm}
  \caption{Self-explaining VLM framework for medical image analysis via report-guided visual attribution. The input image and radiology report are encoded into a shared embedding space. A similarity matrix between image patches and clinical report enables not only diagnosis prediction but also localized visual explanations by highlighting regions semantically aligned with report findings.}
\label{fig:vlm}
\end{figure}

To address this, two key strategies have emerged: post-hoc explainers and intrinsically interpretable architectures. Post-hoc methods augment pretrained VLMs with modules for rationale generation or concept alignment, as in concept-guided prompting for chest X-rays \cite{zhao2024report} and concept bottlenecks integrated with LLMs for skin lesion classification \cite{patricio2024towards}. In contrast, intrinsically interpretable models embed medical concepts directly into the architecture (Fig. \ref{fig:vlm}), exemplified by ConceptCLIP \cite{nie2025conceptclip}, which aligns inputs and concepts in a shared space, and CSR \cite{huy2025interactive}, which uses prototype learning for exemplar-based explanations. Despite these advances, challenges remain in evaluating explanation quality, aligning outputs with clinically validated concepts, and mitigating spurious correlations from pretraining. Future work should prioritize standard benchmarks, task-specific prompting, and concept-aware reasoning frameworks. Table~\ref{tab:vlm_xai_summary} summarizes representative interpretable VLMs and their core strategies.

\begin{table}[htbp]
\caption{Representative interpretable VLMs for biomedical image analysis.}
\label{tab:vlm_xai_summary}
\vspace{-0.2cm}
\resizebox{0.6\textwidth}{!}
{
\begin{tabular}{lll}
\toprule[1pt]
 \textbf{Image Modality} & \textbf{Reference} & \textbf{XAI Method}  \\
    \midrule[1pt]
    CXR + Text &\cite{zhao2024report}      & Concept-guided textual prompting  \\
     CXR + Text &\cite{huy2025interactive}         & CSR (concept + prototype)          \\
    CXR + Text  &\cite{pham2024ai}                & Grad-CAM                           \\
    CXR + Text  &\cite{muller2024chex}                & Visual, textual                          \\
    CXR + Text &\cite{chen2024chexagent}     &Textual   \\
    CXR + Text &\cite{pham2024fg}     &Attention, textual   \\
     Dermatology + Text  & \cite{gao2024aligning}            & Concept bottleneck               \\
     Dermatology + Text &\cite{patricio2024towards}  & CBM + GPT-generated descriptions   \\
     Dermatology + Text &\cite{bie2024xcoop}               & Explainable prompt learning , T-SNE  \\
     Ultrasound, Fluorescence + Text &\cite{wu2025concept}       & Concept bottleneck     \\ 
    \bottomrule[1pt]
\end{tabular}}
\end{table}

\subsection{Summary and Insights}
The integration of XAI into biomedical image analysis reveals modality-specific patterns that shape method selection and interpretability. Visual explanation techniques, particularly CAM-based methods like Grad-CAM, dominate due to their intuitive appeal and compatibility with CNNs. These methods are especially effective in 2D imaging tasks such as chest X-rays and mammography, where localized anomalies are relatively easy to visualize. However, their effectiveness diminishes in more complex settings such as 3D imaging (CT, MRI) or temporal-functional modalities (PET, fMRI), where issues like inconsistent attribution, semantic misalignment, and limited volumetric or temporal coherence arise.

To address these challenges, more sophisticated strategies have been adopted, including perturbation-based methods (e.g., LIME, SHAP), gradient-based attribution (e.g., Integrated Gradients), and concept-based approaches (e.g., ACE, TCAV). These methods provide greater semantic alignment with clinical reasoning and are particularly valuable in high-resolution modalities like histopathology and OCT, where fine-grained structural features carry diagnostic weight. Emerging domains such as ultrasound and nuclear imaging present additional constraints, e.g., operator variability in ultrasound and low resolution in PET/SPECT, which require lightweight, cross-modal, or text-augmented explanation frameworks to improve robustness and clinical relevance.

A key insight is that no single XAI method is universally suitable. Effective interpretability must be modality-aware, task-specific, and user-centered. Hybrid frameworks that combine visual, semantic, and language-based explanations are gaining traction as they better reflect the multifaceted nature of clinical workflows. Moving forward, XAI research should prioritize principled, clinically validated frameworks that balance technical fidelity with human interpretability, grounded in collaboration with domain experts and assessed through real-world clinical utility, not just visual plausibility.

\section{Open Source Frameworks Supporting XAI in Biomedical Image Analysis}\label{sec4} 
The rapid adoption of XAI in biomedical image analysis has been greatly supported by open-source frameworks that implement and standardize interpretability techniques. These tools simplify the integration of XAI into deep learning pipelines and promote reproducibility and benchmarking, both of which are critical for research transparency and clinical translation. In the TensorFlow ecosystem, \texttt{tf-keras-vis}\footnote{\url{https://github.com/keisen/tf-keras-vis}, accessed June 25, 2025} provides a flexible interface for generating saliency maps, activation maximization, and Grad-CAM variants, and is commonly used in convolutional models for classification and segmentation tasks.
In the PyTorch ecosystem, several mature libraries have emerged. \texttt{Captum}\footnote{\url{https://github.com/pytorch/captum}, accessed June 25, 2025}, developed by Meta AI, supports a wide range of attribution methods—including Integrated Gradients, DeepLIFT, SHAP, and TCAV—and is well-suited for multimodal biomedical data. \texttt{TorchRay}\footnote{\url{https://github.com/facebookresearch/TorchRay}, accessed June 25, 2025} offers tools for visual attribution and counterfactual analysis, while \texttt{pytorch-grad-cam}\footnote{\url{https://github.com/jacobgil/pytorch-grad-cam}, accessed June 25, 2025} and \texttt{TorchCAM}\footnote{\url{https://github.com/frgfm/torch-cam}, accessed June 25, 2025} provide lightweight support for CAM-based methods, frequently used in radiology, pathology, and ophthalmology.
Together, these modular and well-documented frameworks lower the barrier to entry and will continue to play a key role in translating XAI from research to clinical application. Table~\ref{tab:framework} summarizes widely used XAI toolkits, including their supported methods and modality compatibility.

\begin{table}[htbp] 
\caption{Representative open-source XAI frameworks for biomedical image analysis: supported methods, backends, and access links.}
\label{tab:framework}
\vspace{-0.2cm}
\resizebox{0.85\textwidth}{!}
{
\begin{tabular}{p{2.5cm} p{6cm} p{3cm} p{3.5cm}}
\toprule[1pt]
\textbf{Framework} & \textbf{Supported XAI Methods} & \textbf{Supported Backend} & \textbf{Access Link} \\
\midrule[1pt]
tf-keras-vis & 
Vanilla saliency \cite{simonyan2013deep}, 
Smooth-Grad \cite{smilkov2017smoothgrad},\newline 
Grad-CAM \cite{8237336}, 
Grad-CAM++ \cite{8354201}, \newline
Score-CAM \cite{9150840}, 
Faster-ScoreCAM\footnote{\url{https://github.com/tabayashi0117/Score-CAM\#faster-score-cam}, accessed June 25, 2025}, \newline
Layer-CAM \cite{9462463} 
& TensorFlow, Keras & 
\url{https://github.com/keisen/tf-keras-vis?tab=readme-ov-file} \\
\hline

pytorch-grad-cam & 
Grad-CAM \cite{8237336} ,
Eigen-CAM \cite{Muhammad_2020}, \newline
Grad-CAM++ \cite{8354201},
HiResCAM \cite{draelos2021use}, \newline
FullGrad \cite{srinivas2019fullgradient}, 
XGrad-CAM \cite{fu2020axiombased}, \newline
Ablation-CAM \cite{9093360},
Score-CAM \cite{9150840}, \newline
Eigen Grad-CAM\footnote{\url{https://github.com/jacobgil/pytorch-grad-cam}, accessed June 25, 2025},
Layer-CAM \cite{9462463}, \newline
Deep Feature Factorizations \cite{collins2018deep} 
& PyTorch & 
\url{https://github.com/jacobgil/pytorch-grad-cam} \\
\hline

CAPTUM & 
SmoothGrad \cite{smilkov2017smoothgrad}, 
DeConvNet \cite{zeiler2014visualizing}, \newline
Guided-BackProp \cite{springenberg2014striving},
LRP \cite{Bach2015}, \newline
DeepLIFT \cite{shrikumar2017learning}, 
LIME \cite{ribeiro2016should},
SHAP \cite{lundberg2017unified},\newline
IG \cite{sundararajan2017axiomatic}, 
TCAV \cite{kim2018interpretability},
Occlusion \cite{zeiler2014visualizing} 
& PyTorch & 
\url{https://captum.ai/tutorials/TorchVision_Interpret/} \\
\hline

TorchRay & 
DeConvNet~\cite{zeiler2014visualizing}, Grad-CAM \cite{8237336},\newline
Guided-BackProp~\cite{springenberg2014striving}, RISE \cite{petsiuk2018rise} 
& PyTorch & 
\url{https://facebookresearch.github.io/TorchRay/} \\
\hline

TorchCam  & 
CAM \cite{7780688}, 
SS-CAM \cite{wang2020sscam}, 
IS-CAM \cite{naidu2020iscam},\newline
Grad-CAM \cite{8237336}, 
Grad-CAM++ \cite{8354201}, \newline
Smooth Grad-CAM++ \cite{omeiza2019smooth}, 
Score-CAM \cite{9150840}, \newline
XGrad-CAM \cite{fu2020axiombased},
Layer-CAM \cite{9462463} 
& PyTorch & 
\url{https://frgfm.github.io/torch-cam/index.html} \\
\bottomrule[1pt]
\end{tabular} }
\end{table}

\section{Evaluation Metrics for XAI in Biomedical Image Analysis} \label{sec5}
As XAI becomes more prevalent in biomedical image analysis, evaluating the quality and utility of explanations is essential. Unlike traditional metrics such as accuracy or AUC, XAI evaluation must consider how well explanations align with human reasoning, reflect actual model behavior, and support clinical decisions, often without a clear ground truth for correctness \cite{hossain2025explainable}. In this section, we review key evaluation metrics tailored for biomedical imaging, examining their assumptions, applicability to different modalities and tasks, and limitations in clinical settings.

\subsection{Evaluation Metrics for Visual Explanations}

\textbf{Relevance Mass Accuracy (RMC)} and \textbf{Relevance Rank Accuracy (RRA)} are two widely used metrics to evaluate how well explanation heatmaps align spatially with annotated clinical regions~\cite{Arras_2022}. RMC measures the proportion of total relevance concentrated within a ground truth mask (e.g., tumor or organ region), computed as:
\begin{math}
     \text{RMC} = \frac{\sum_{p\in {GT}}R_{p}}{\sum_{p\in \text{image}}R_{p}},
\end{math}
where $R_{p}$ is the relevance at pixel $p$ in the heatmap, and $GT$ is the set of pixels within the annotated region. A higher RMC indicates stronger spatial alignment between model attention and clinically important areas. RRA focuses on rank-based localization. It evaluates whether the most relevant pixels (top-$K$, where $K=|GT|$)coincide with the ground truth: $\text{RRA} = \frac{|\text{Top-}k \cap GT|}{|GT|}.$
This captures the explanation's ability to prioritize the correct regions among the most influential pixels. Both metrics are especially useful in biomedical imaging tasks like lesion detection or anatomical structure segmentation, where spatial fidelity is critical to clinical interpretability.

\textbf{Deletion} and \textbf{Insertion} are fidelity-based metrics that evaluate how well a saliency map reflects the model's decision-making process \cite{samek2015evaluating}. Deletion measures the drop in class confidence as the most relevant pixels, identified by the explanation, are progressively removed from the input. In contrast, insertion assesses the increase in confidence as these pixels are gradually added to a blank or blurred baseline. Steeper confidence curves in both metrics indicate higher explanation fidelity, suggesting the highlighted regions are truly influential. These metrics are particularly valuable in biomedical imaging as they can verify the causal relevance of identified features (e.g., lesions or anatomical structures).

\textbf{Pointing Game} evaluates the spatial accuracy of heatmaps by checking whether the most activated point falls within the ground truth region \cite{zhang2016topdown}. If the peak relevance lies inside the annotated area, it is counted as a \textit{hit}; otherwise, as a \textit{miss}. The localization accuracy is then defined as:
\begin{math}  
\text{Accuracy} = \frac{\text{Number of \textit{Hits}}}{\text{Number of \textit{Hits}} + \text{Number of \textit{Misses}}}. 
\end{math}
This metric offers a simple yet effective way to evaluate whether the model’s focus aligns with clinically relevant areas. Unlike pixel-wise overlap metrics, it requires only point-level correspondence rather than full segmentation masks and is less sensitive to annotation boundaries, making it practical for varied biomedical imaging tasks.

\textbf{Area Over Perturbation Curve (AOPC)} evaluates the fidelity of heatmaps by measuring how the model's confidence declines as the most relevant regions are progressively occluded \cite{samek2016evaluating}. Given an image $x$ and a ranked heatmap, top-ranked image regions are occluded $K$ steps, and AOPC quantifies the average confidence drop in the model output $f(x)$ over $K$ steps:
   $ \text{AOPC} = \frac{1}{N} \sum_{n=1}^{N} \left( f(x) - \frac{1}{K} \sum_{k=1}^{K} f(x^{k}) \right), $
where $x^{k}$ is the $k$-th perturbed image. In biomedical image analysis, a higher AOPC suggests that the identified regions indeed contribute meaningfully to the model’s decision.

\subsection{Evaluation Metrics for Non-Visual Explanation}

\textbf{Bilingual Evaluation Understudy (BLEU)} is a widely used metric for evaluating automatically generated textual explanations by measuring their $n$-gram overlap with reference texts \cite{papineni2002bleu}. It combines modified $n$-gram precision with a brevity penalty to penalize overly short outputs. The BLEU score is defined as: 
\begin{math}
     \text{BLEU} = BP \cdot \exp\left( \sum_{n=1}^{N} w_n \log p_n \right)
\end{math}
where $BP$ is the brevity penalty, $w_n$ is the weight for each $n$-gram level, and $p_n$ is the precision of modified $n$-grams. Score ranges from 0 to 1, with higher values indicating closer alignment with reference texts. In biomedical image analysis, BLEU has been applied to  assess the quality of generated reports or captions (e.g., radiology or ultrasound), though it primarily captures surface-level similarity and may not reflect semantic fidelity or clinical adequacy \cite{alsharid2019captioning,gajbhiye2020automatic}.

\textbf{Metric for Evaluation of Translation with Explicit Ordering (METEOR)} is a reference-based metric that evaluates the quality of XAI-generated text by aligning it with expert-written references \cite{banerjee2004meteor}. Unlike BLEU, which emphasizes precision, METEOR balances both precision and recall and accounts for word order, stemming, synonyms, and paraphrases, making it more robust to linguistic variation. The score is computed as:
\begin{math}
     \text{METEOR} = F_\text{mean} \cdot (1 - \text{Penalty}),
\end{math}
where $F_\text{mean}$ is the harmonic mean of unigram precision and recall, and the penalty reflects alignment fragmentation. In biomedical image analysis, METEOR has been used to assess the fluency and content fidelity of generated reports or rationales, offering higher alignment with human judgment than purely lexical metrics.

\textbf{Recall-Oriented Understudy for Gisting Evaluation (ROUGE)} is a set of reference-based metrics for evaluating XAI-generated text by measuring lexical overlap with human-authored descriptions \cite{lin-2004-rouge}. Among its variants, ROUGE-L is well-suited for biomedical applications, as it captures sentence-level structure through the Longest Common Subsequence (LCS) between candidate and reference texts. The ROUGE-L F1 score is defined as:
\begin{math}
    \text{ROUGE-L}_{F1} = \frac{(1+\beta^2)\cdot \text{LCS}(c,g)}{m+ \beta^2\cdot n},
\end{math}
where $\text{LCS}(c,g)$ is the length of the longest common subsequence between the candidate caption $c$ and reference $g$, and $m$, $n$ are their respective lengths. The parameter $\beta$ adjusts the relative weight of recall versus precision. Unlike $n$-gram-based metrics, ROUGE-L does not require consecutive word matches, making it effective for evaluating fluency and coherence in clinical reports, diagnostic rationales, and other structured textual outputs in XAI.

\textbf{Consensus-Based Image Description Evaluation (CIDEr)} is a consensus-based metric that evaluates XAI-generated captions by comparing them to a set of expert-written references using weighted $n$-gram similarity \cite{vedantam2015cider}. Each $n$-gram is encoded as a term frequency-inverse document frequency (TF-IDF) vector to emphasize informative phrases and downweight common ones. The CIDEr score is computed as the average cosine similarity between the candidate caption and all references across $n$-gram levels:
\begin{math}
    \text{CIDEr}_n(c, S) = \frac{1}{N} \sum_{n=1}^N \frac{1}{|S|} \sum_{s \in S} \frac{g^n(c) \cdot g^n(s)}{\|g^n(c)\| \, \|g^n(s)\|},
\end{math}
where $g^n(\cdot)$ is the TF-IDF vector of $n$-grams, $c$ is the candidate caption, and $S$ is the reference set. Compared to metrics rely on exact $n$-gram matches, CIDEr captures both syntactic fluency and semantic relevance, making it particularly suited for assessing long-form clinical explanations, such as radiology report generation and multimodal diagnostic summaries.

\textbf{Semantic Propositional Image Caption Evaluation (SPICE)} assesses the semantic quality of generated captions by converting them into scene graphs that capture objects, attributes, and relationships \cite{anderson2016spice}. It compares these structured semantic tuples between the candidate and reference texts, focusing on meaning rather than surface-level $n$-gram matches. The score is computed as the F1-measure between matched tuples:
\begin{math}
    \text{SPICE} = \frac{2\cdot|\text{Matches}|}{|\text{Candidate Tuples}|+|\text{Reference Tuples}|}.
\end{math}
By emphasizing semantic propositions, SPICE is particularly well-suited for evaluating  XAI-generated medical report, where accurate representation of clinical entities and their relationships is critical for interpretability.

\section{Open Challenges and Future Directions} \label{sec6}
\subsection{Current Limitations of XAI in Biomedical Image Analysis}
Despite increasing interest, XAI in biomedical image analysis faces critical limitations that restrict its clinical applicability and reliability. A major issue is the lack of modality-aware design. Most XAI techniques are adapted from general computer vision tasks and do not account for the unique spatial, anatomical, or resolution properties of biomedical modalities. For example, heatmap-based methods such as Grad-CAM remain popular despite their limited spatial precision in volumetric imaging (e.g., MRI, CT) and weak alignment with tissue-level structures in histopathology or ultrasound. Furthermore, integration into real-world workflows is rare. High computational cost, lack of intuitive visualization interfaces, and limited clinician training in interpreting model outputs all contribute to a gap between research development and practical deployment.

Another persistent limitation is the weak alignment of model explanations with human semantics and clinical reasoning. Many saliency-based methods highlight low-level features without clarifying their diagnostic relevance, while concept-based and textual methods require costly annotation or risk generating oversimplified rationales. Compounding this is the lack of standardized evaluation protocols. Current evaluation metrics, such as deletion, insertion, or hit-rate, are inconsistently applied and often fail to capture clinical utility or reasoning processes. Without benchmarks tailored to biomedical image tasks, reproducibility and cross-study comparison remain challenging. Addressing these gaps requires not only methodological innovation but also closer alignment with clinical expectations and diagnostic workflows.

\subsection{Open Challenges and Future Directions}
Despite increasing interest and progress in XAI for biomedical image analysis, several key challenges remain unresolved. This section outlines five forward-looking directions that address current limitations and guide future research.

\textbf{1. Modality- and Task-Aware Interpretability.} Biomedical imaging spans modalities such as CT, MRI, PET, ultrasound, and histopathology, each with distinct signal characteristics and diagnostic goals. However, most XAI techniques remain modality-agnostic and task-invariant. Future work should incorporate modality-specific priors, spatial constraints, and acquisition-aware features. Likewise, explanations should adapt to task types, such as classification, segmentation, or treatment planning, by aligning with the decision-making processes relevant to each.

\textbf{2. Semantically Grounded and Clinically Meaningful Explanations.} Current XAI outputs often lack semantic alignment with clinical reasoning. Future models must integrate domain knowledge to generate human-understandable explanations. Self-supervised concept discovery, ontology-guided attribution, and alignment with clinical documentation (e.g., EMRs or radiology reports) can help ground visual and textual outputs in medical semantics. The shift from saliency to structured, interpretable rationale is key to clinical acceptance.

\textbf{3. Reliable and Standardized Evaluation Frameworks.} Evaluating XAI remains inconsistent and fragmented. Existing metrics do not always capture clinical relevance. Future directions should establish domain-specific, task-grounded, and standardized evaluation protocols. Models with embedded interpretability, optimized during training, can facilitate more consistent assessments. Additionally, hybrid evaluations that combine human expert feedback with fidelity and robustness benchmarks will yield a fuller picture of explanation quality.

\textbf{4. Clinically Usable and Adaptive Systems.} Most XAI models are not designed with clinical usability in mind. Future systems should support context-aware explanation granularity, tailored to user roles (e.g., radiologist vs technician) and task demands (e.g., triage vs diagnosis). Human factors such as cognitive load and decision context must inform interface design. Interactive, human-in-the-loop systems that adapt based on user feedback could bridge the gap between research and clinical practice.

\textbf{5. Generalizable, Modular, and Transparent XAI Architectures.} Current systems lack scalability and cross-domain generalization. Future XAI frameworks should unify visual, conceptual, and textual explanations through modular design. Emphasizing explanation provenance, reproducibility, and training-time interpretability will be essential for regulatory compliance and clinical trust. Modular components, such as saliency engines, concept mappers, or captioning modules, should be reusable across tasks and settings, accelerating deployment and standardization.

\section{Conclusion} \label{sec7}
This survey provided a comprehensive and modality-aware overview of XAI techniques in biomedical image analysis. We systematically categorized existing methods, analyzed their foundations and limitations, and introduced a taxonomy that maps XAI approaches to specific imaging modalities. We also reviewed recent developments in multimodal learning and vision-language models, expanding the scope of explainability beyond visual attributions. In addition, we summarized key evaluation metrics and open-source toolkits that support implementation and benchmarking. Despite recent advances, challenges remain in modality-specific design, semantic alignment, evaluation standardization, clinical usability, and scalability. We identified these limitations and outlined future research directions to guide the development of more trustworthy and clinically meaningful XAI systems. By combining technical depth with practical insight, this work offers a structured reference for advancing interpretable deep learning in biomedical imaging.

\bibliographystyle{ACM-Reference-Format}
\bibliography{sample-base}

\appendix

\end{document}